\documentclass[format=acmsmall, review=false, screen=true]{acmart}

\usepackage[lined,ruled,linesnumbered]{algorithm2e}

\SetAlFnt{\small}
\SetAlCapFnt{\small}
\SetAlCapNameFnt{\small}
\SetAlCapHSkip{0pt}
\IncMargin{-\parindent}

\acmJournal{JOCCH}

\received{September 2021}

\newif\ifdraft
\drafttrue

\usepackage{amsmath}
\usepackage{array}
\usepackage[english]{babel}
\usepackage{color}
\usepackage{colortbl}
\usepackage{dcolumn}
\usepackage{graphicx}
\usepackage[utf8]{inputenc}
\usepackage{latexsym}
\usepackage{lineno}
\usepackage{multibib}
\usepackage{multicol}
\usepackage{multirow}
\usepackage{pifont}
\usepackage{rotating}
\usepackage{soul}
\usepackage{url}

\usepackage{multibib}

\newcites{prim}{Primary sources}
\newcites{sec}{References}

\newcommand{\blue}[1]{\ifdraft{\leavevmode\color{black}{#1}}\else{\leavevmode\color{black}{#1}}\fi}
\newcommand{\wasblue}[1]{\ifdraft{\leavevmode\color{black}{#1}}\else{\leavevmode\color{black}{#1}}\fi}

\newcommand{\tfidf}{\mathrm{tfidf}}

\newcommand{\TP}{\mathrm{TP}}
\newcommand{\FP}{\mathrm{FP}}
\newcommand{\FN}{\mathrm{FN}}
\newcommand{\TN}{\mathrm{TN}}

\newtheorem{revcomment}{Reviewer Comment}[section]

\newcolumntype{M}[1]{>{\centering\arraybackslash}p{#1}}
\newcolumntype{L}{>{\centering\arraybackslash}m{12cm}}

\usepackage[normalem]{ulem}
\usepackage{nicefrac}

\newcommand{\medone}{\textsc{MedLatinEpi}}
\newcommand{\medtwo}{\textsc{MedLatinLit}}
\newcommand{\medievalla}{\textsc{MedieValla}}

\newcommand{\side}[1]{\begin{sideways}{#1}\end{sideways}}

\newcommand{\killpunct}[1]{}

\newcolumntype{L}{>{\arraybackslash}m{23em}}
\newcolumntype{R}{>{\arraybackslash}m{8em}}

\begin{document}

\title{\medone\ and \medtwo: Two Datasets for the Computational
Authorship Analysis of Medieval Latin Texts}
\thanks{The order in which the authors are listed is purely
alphabetical; each author has given an equally important contribution
to this work.}

%

\author{Silvia Corbara} \orcid{0000-0002-5284-1771} \affiliation{%
\institution{Scuola Normale Superiore} \postcode{56126} \city{Pisa}
\country{Italy} } \email{silvia.corbara@sns.it}

\author{Alejandro Moreo} \orcid{0000-0002-0377-1025} \affiliation{
\department{Istituto di Scienza e Tecnologie dell'Informazione}
\institution{Consiglio Nazionale delle Ricerche} \postcode{56124}
\city{Pisa} \country{Italy} } \email{alejandro.moreo@isti.cnr.it}

\author{Fabrizio Sebastiani}\orcid{0000-0003-4221-6427} \affiliation{
\department{Istituto di Scienza e Tecnologie dell'Informazione}
\institution{Consiglio Nazionale delle Ricerche} \postcode{56124}
\city{Pisa} \country{Italy} } \email{fabrizio.sebastiani@isti.cnr.it}

\author{Mirko Tavoni} \orcid{0000-0002-3063-1503} \affiliation{%
\department{Dipartimento di Filologia, Letteratura e Linguistica}
\institution{Università di Pisa} \postcode{56126} \city{Pisa}
\country{Italy} } \email{mirko.tavoni@unipi.it}

\renewcommand{\shortauthors}{Corbara, Moreo, Sebastiani, Tavoni}

\renewcommand{\shorttitle}{Two Datasets for the Computational
Authorship Analysis of Medieval Latin Texts}

\begin{abstract}
  We present and make available \wasblue{\medone} and
  \wasblue{\medtwo}, two datasets of medieval Latin texts to be used
  in research on computational authorship analysis. \medone\ and
  \medtwo\ consist of 294 and 30 curated texts, respectively, labelled
  by author; \medone\ texts are of epistolary nature, while \medtwo\
  texts consist of literary comments and treatises about various
  subjects. As such, these two datasets lend themselves to supporting
  research in authorship analysis tasks, such as authorship
  attribution, authorship verification, or same-author
  verification. \wasblue{Along with the datasets we provide
  experimental results, obtained on these datasets, for the authorship
  verification task, i.e., the task of predicting whether a text of
  unknown authorship was written by a candidate author or not. We also
  make available the source code of the authorship verification system
  we have used, thus allowing our experiments to be reproduced, and to
  be used as baselines, by other researchers. We also describe the
  application of the above authorship verification system, using these
  datasets as training data, for investigating the authorship of two
  medieval epistles whose authorship has been disputed by scholars.}
 
\end{abstract}

\keywords{Authorship Analysis, Authorship Verification, Medieval
Latin}

\maketitle


\section{Introduction}
\label{sec:intro}

\noindent (Computational) \emph{Authorship Analysis} is the task of
inferring the characteristics of the author of a text of unknown or
disputed authorship. Authorship Analysis has several subtasks of
practical use; examples include \emph{gender detection} (i.e.,
predicting whether the text was written by a woman or a man
\citesec{Koppel02}), or \emph{native language identification} (i.e.,
predicting the native language of the author of the
text~\citesec{Perkins:2015rp}).

Many subtasks of authorship analysis have actually to do with the
prediction of the \emph{identity} of the author of the text. The one
such subtask that has the longest history is \emph{Authorship
Attribution}
(AA)~\citesec{Juola:2006fk,Koppel:2009ix,Stamatatos2009:yq}, which
consists of predicting who, among a set of $k$ candidate authors, is
the real \wasblue{(or the most likely)} author of the text. A task
that has gained prominence more recently is \emph{Authorship
Verification} (AV)~\citesec{Koppel:2004ja,Stamatatos:2016ij}, the task
of predicting if a certain candidate author is or is not the author of
the text. Finally, the task that has been introduced latest in this
field is \emph{Same-Authorship Verification}
(SAV)~\citesec{Koppel:2014bq}, the task of predicting whether two
texts $d'$ and $d''$ are by the same author or not.

Nowadays, authorship analysis tasks are usually tackled as \emph{text
classification} tasks~\citesec{Aggarwal:2012wl}, and thus solved with
the help of machine learning algorithms: for instance, an authorship
verification task is solved as a \emph{binary classification} problem,
i.e., as the problem of classifying the disputed text into one of the
two classes \{\textsc{Yes}, \textsc{No}\}, where \textsc{Yes} (resp.,
\textsc{No}) indicates that the text is (resp., is not) by the
candidate author. In order to do so, a machine learning algorithm
trains a \{\textsc{Yes}, \textsc{No}\} classifier from a training set
of labelled texts, where the training examples labelled \textsc{Yes}
are texts by the candidate author and the training examples labelled
\textsc{No} are texts by other authors, usually closely related to the
candidate author.

Authorship analysis is useful for many applications, ranging from
cybersecurity (the field that addresses the design of techniques for
preventing crimes committed via digital
means)~\citesec{Schmid:2015qq}, to computational forensics (the field
concerned with the study of digital evidence for investigating crimes
that have already
occurred)~\citesec{Chaski:2005pd,Larner:2014kl,Perkins:2015rp,Rocha:2017yy}.
Another important application is related to philology, and has to do
with inferring the identity of the unknown authors of texts of
literary and historical value. In the case of modern texts, this often
has to do with the attempt to disclose the identity of authors who
originally wanted to remain anonymous, \wasblue{or to disguise as
someone else}, while in the case of ancient texts this usually has to
do with texts whose authorship has \textit{become} unknown, or
uncertain, in the course of
history~\citesec{Kabala:2020bu,Kestemont:2016fh,Savoy:2019qr,Stover:2016zl,Tuccinardi:2017yg}.

\wasblue{After reviewing some related work
(Section~\ref{sec:relatedwork}),} we here present and make available
(Section~\ref{sec:datasets}) two datasets of texts of the latter type,
i.e., texts written in medieval Latin, mostly by Italian
literates, mostly dating around the 13th and 14th
century.\footnote{Medieval Latin is different from classical Latin in
a number of ways, e.g., it is more generous than classical Latin in
its use of prepositions and conjunctions, and it uses a more regular
syntax.} We believe this to be an important contribution for at least
two reasons. The first is that the datasets bring together (in
preprocessed form for use by authorship analysis researchers) a set of
texts that were not readily available to these researchers, since some
of these texts were not available in digital form, while others lay
scattered across different electronic formats and different digital
libraries. The second is that there are many documents in medieval
Latin from this historical period whose authorship is disputed by
scholars,\footnote{\label{foot:pellegrini}Examples include the
\emph{Epistle to Cangrande}~\citesec{Casadei:2020tm},
\emph{Cangrande's Epistle to Henry VII}~\citesec{Pellegrini:2018yb},
and the \emph{Quaestio de aqua et terra}~\citesec{Toynbee:1918mj},
just to mention ones that some scholars attribute to Dante Alighieri
while some others do not. \wasblue{The first two will be discussed
more fully in Section~\ref{sec:twocasestudies}.}} and this makes an
authorship analysis system trained on these datasets an important tool
for philologists and historians of language alike.


Aside from describing the two datasets, we make available the source
code of \medievalla,\footnote{The name \medievalla\ is a combination
of ``medieval'' and the last name of Lorenzo Valla (1407--1457), one
of the first (human) authorship verifiers recorded in history. Lorenzo
Valla is well-known for proving that the so-called ``Donation of
Constantine'' (a decree attributed to 4th-century Roman emperor
Constantine in which he supposedly conferred authority over Rome and
the western part of the Roman Empire to the Pope) was a forgery.}
a software tool for running authorship verification experiments on
medieval Latin texts, and we present (Section~\ref{sec:results}) the
results of our experiments using \medievalla\ on these datasets. The
availability of both the datasets and the tool we have used on them,
will allow other researchers to replicate our results and, hopefully,
to develop and test improved authorship verification methods for
medieval Latin.

\wasblue{In Section~\ref{sec:twocasestudies} we present two example
applications of \medievalla\ on \medone\ and \medtwo. In the first we
verify if the \textit{Epistle to Cangrande}, an epistle traditionally
attributed to Dante Alighieri, but which several scholars have
conjectured to be a forgery, is actually by Dante or not. In the
second we verify if an epistle traditionally attributed to Cangrande
della Scala, but which has recently been conjectured to be by Dante
himself, is indeed by Dante or not. In both cases our authorship
verifier rejects the hypothesis that Dante may be the author, and does
so with high confidence.}


\section{Related work}
\label{sec:relatedwork}

%
\noindent In our cultural heritage, documents of unknown or disputed
authorship are rather common, especially in centuries-old traditions,
where the testimony of the true author may easily have been lost or
altered. In particular, a number of recent works have tackled problems
of authorship analysis for the Latin language.

In \citesec{Kestemont:2015lp}, Kestemont et al.\ address an authorship
attribution task characterised by two disputed documents written in
medieval Latin, and three possible authors -- the well-known Christian
mystic Hildegard of Bingen, her secretary Guibert of Gembloux, and
Bernard of Clairvaux. They employ a PCA-based approach
on the frequencies of $65$ function words.
%
In a later work \citesec{Kestemont:2016fh}, Kestemont et al.\ tackle
another authorship attribution task concerning
parts of the \textit{Corpus Caesarianum}, including in the candidate
set Caesar, his general Aulus Hirtius, and three other unidentified
authors.
The methodologies they adopt is based on comparing an author's profile
(where an author's profile is defined as the centroid of the vectors
corresponding to that author's known texts) with the document of
disputed authorship. Two different techniques are employed in this
work, namely, the distance between the vectors representing the
author's profile and the disputed document, and a generic
implementation of Koppel and Winter's ``impostors
method''~\citesec{Koppel:2014bq}. With both techniques the authors use
word unigrams and character $n$-grams as features, and test their
systems on the datasets from the Authorship Verification track at
PAN2014 and on a corpus of historic Latin authors.
They perform experiments with various distance metrics and vector
space models for both techniques.

An approach that similarly exploits the concept of author profile can
be found in~\citesec{Tuccinardi:2017yg}, a study regarding the
authenticity of one of Pliny the Younger’s letters. In particular, the
author employs the ``simplified profile intersection'', a similarity
measure that uses the size of the intersection among the profile of
the unknown document and that of the target author's production, which
is computed by counting the $n$-grams in common between them. In order
to find the model with the best discriminating power between Pliny's
and non-Pliny's writing, additional fragments of letters from Cicero
and Seneca are employed.

Stover et al.'s \citesec{Stover:2016zl} is yet another work that
employs Koppel and Winter's ``impostors
method''~\citesec{Koppel:2014bq}. Here, a newly found Latin document
is investigated in a same-authorship verification setting, where word
unigrams and bigrams are used as features. Ultimately, the only
textual pair that receives a satisfying positive score is the one
consisting of the disputed document and \textit{De Platone} by
Apuleius, hence strongly supporting the hypothesis that Apuleius may
be the author of the document.

Vainio et al.'s \citesec{Vainio:2019cg} is the only study, among the
ones we consider here, that uses a deep-learning algorithm.  In
particular, the authors train both an SVM and a CNN for an authorship
verification task, consisting of recognising Cicero's written style
against the styles of the background authors, and then use the two
trained classifiers to classify four disputed documents.  They conduct
various experiments with POS-grams and character $5$-grams. This
dataset, which is freely downloadable, is the one with the largest
number of authors among the works discussed in this section, counting
$44$ authors, including anonymous and pseudo-authors; this is thanks
to the wide timeframe considered, which goes from the 1st century BC
to the 5th century AD.

Kabala's~\citesec{Kabala:2020bu} is the only work that, like the
present paper, focuses on \textit{medieval} Latin, although dating
from an earlier period than the one we consider. It performs
same-authorship verification on two texts, the \textit{Translatio s.\
Nicolai} and the \textit{Gesta principum polonorum}. In particular,
through the studies on four different datasets, the author seeks to
understand whether the alleged authors of the two documents, the
so-called Monk of Lido and Gallus Anonymous, are actually the same
person.
The study is conducted by classifying both texts with respect to the
author classes within each dataset, using $9$ distance metrics and
logistic regression. Each dataset counts between $39$ and $116$ texts
dating from the 10th to the 12th centuries, written by between $15$
and $22$ different authors. These are the only datasets of medieval
Latin texts that are freely available to the public among the ones we
have surveyed in this section.

While the above works focus on cases of uncertain paternity, such
methodologies might also be applied to documents of certain
authorship, e.g., in order to study possible stylistic influences
among authors. In Forstall et al.'s work~\citesec{Forstall:2011ev},
for example, the goal is to verify a supposed influence by Catullo on
the poetry of Paul the Deacon. Forstall et al.'s idea is to train an
SVM with samples of Catullo's writings (in a typical authorship
verification setting), employing various kinds of $n$-grams as
features. A document highly influenced by Catullo, thus bearing many
similarities to his style, will then receive a high classification
score by the AV system.

In Table~\ref{tab:related_works} we summarise the works discussed in
this section, specifying the task being tackled, the number of authors
in the dataset, the method of analysis and the features employed, the
dataset sources, and whether the dataset is publicly available or not.

\begin{table}[t]
  \caption{Main characteristics of published works on authorship
  analysis for the Latin language reviewed in
  Section~\ref{sec:relatedwork}. ``AA'' stands for authorship
  attribution, ``AV'' stands for authorship verification, while
  ``SAV'' stands for same-author verification. The works are in
  alphabetical order by first author.}
  \begin{center}
    \resizebox{\textwidth}{!} {
    \begin{tabular}{|l||c|c|p{4cm}|p{4cm}|p{4cm}|c|}
      \hline 
      & \multicolumn{1}{c|}{\side{Task}} & \multicolumn{1}{c|}{\side{Number of authors in dataset\phantom{0}}} & 
                                                                                                                 \multicolumn{1}{c|}{\side{Methods used}} & \multicolumn{1}{c|}{\side{Features used}} & \multicolumn{1}{c|}{\side{Dataset}} & 
                                                                                                                                                                                                                                              \multicolumn{1}{c|}{\side{Makes the dataset available}} \\
      \hline \hline 
      Forstall et al.\ \citesec{Forstall:2011ev}   &  AV & 7 & SVM &  Functional $n$-grams (on text and metric) and low-probability $n$-grams & Transcriptions and Tesserae & No\\
      \hline 
      Kabala \citesec{Kabala:2020bu}     & SAV & 15--22 & Distance metrics and Logistic regression & 250 most frequent words & Patrologia Latina and Latin Library & Yes \\
      \hline  
      Kestemont et al.\ \citesec{Kestemont:2015lp}  & AA & 3 & PCA & 65 function words & Brepols Publishers & No\\
      \hline 
      Kestemont et al.\ \citesec{Kestemont:2016fh}  & AA & \vtop{\hbox{\strut29 (dev)}\hbox{\strut~3 (test)}} & Distance metrics on author's profile and Impostors method & Word unigrams and char $n$-grams & Latin Library & Yes \\
      \hline 
      Stover et al.\ \citesec{Stover:2016zl}     & SAV & 36 & Impostors method & Word unigrams and bigrams & Brepols Publishers and Latin Library and Patrologia Latina & Partially \\
      \hline 
      Tuccinardi \citesec{Tuccinardi:2017yg} & AV & 3 & Simplified Profile Intersection & Character $n$-grams & [unspecified] & No \\
      \hline 
      Vainio et al.\ \citesec{Vainio:2019cg}     & AV & 44 & SVM and CNN & Pos-tags, word and char $n$-grams & Latin Library and Bibliotheca Augustana & Yes \\
      \hline 
    \end{tabular}
    }
  \end{center}
  \label{tab:related_works}
\end{table}%

In general, it should be noted that the authors do not subject
their datasets to a thorough cleaning from information extraneous to
the author's production. In particular, citations of other authors
(i.e., pieces of text that are by someone other than the author of the
citing text) are seldom removed (in some cases, only the most
extensive ones are); this may hamper authorship analysis, since cited
text ``contaminates'' the citing text, at least as far as authorship
analysis is concerned. This is unlike the present paper, where cited
text is scrupulously removed.


\section{The datasets}
\label{sec:datasets}


\subsection{Origin of the datasets}
\label{sec:origin}

\noindent Our two datasets originated in the context of an authorship
verification research work~\citesec{Corbara:2019cq,Corbara:2019rw}
that we carried out in order to establish, using an approach based on
machine learning, whether the \textit{Epistle to Cangrande},
originally attributed to Dante Alighieri, is actually a forgery or
not, a fact which is intensely debated among philologists
today~\citesec{Casadei:2020tm}. The \textit{Epistle to Cangrande} is
traditionally listed as the 13th of Dante's epistles that have reached
us; hereafter we will thus refer to it as Ep13.

Ep13 is written in medieval Latin and addressed to Cangrande I, ruler
of the Italian cities of Verona and Vicenza at the beginning of the
14th century. Scholars traditionally divide it into two portions that
are distinct in purpose and, consequently, style: the first portion
(paragraphs 1--13, hereafter: Ep13(I)) is the dedicatory section, with
proper epistolary characteristics, while the second portion
(paragraphs 14--90, hereafter: Ep13(II)) contains an exegesis (i.e.,
analysis) of Alighieri's \textit{Divine Comedy}, and in particular a
commentary of the first few lines of its third part, the
\textit{Paradise}. Scholars are not unanimous as whether Dante
Alighieri is the true author of Ep13: some of them consider both
portions authentic, some consider both portions the work of a forger,
while others consider the first part authentic and the other a
forgery.

Since it is unclear whether the two portions are by the same author or
not, we tackled our AV problem as two separate AV sub-problems, one
for Ep13(I) and one for Ep13(II). Because of the different nature of
the two portions, we built two separate training sets, one for Ep13(I)
and one for Ep13(II); we will refer to them as \medone\
\wasblue{(where ``\textsc{Epi}'' refers to the \underline{epi}stolary
nature of the texts contained therein)} and \medtwo\ \wasblue{(where
``\textsc{Lit}'' stands for \underline{lit}erary)}, respectively.

In both \medone\ and \medtwo\ Dante Alighieri is, of course, the
author of some of the labelled texts. The texts attributed with
certainty to Alighieri and written in Latin are few and well known; we
have thus included all of them.\footnote{We have not included the
\textit{Quaestio de aqua et terra}, a work traditionally attributed to
Dante Alighieri, exactly because its authorship is currently
disputed. Other works by Alighieri, such as his masterpiece
\emph{Divina Commedia}, are not included because they are written not
in Latin but in the Florentine \wasblue{vernacular}, the language that
would later form the basis of the Italian language.} Concerning other
authors, the approach we have chosen is to select literates who are as
``close'' (culturally and stylistically) to Dante Alighieri as
possible, i.e., authors whose production is characterised by
linguistic features similar to Alighieri's. The reason for this
choice, of course, is that, if the non-Dantean texts used for training
were very different from Dante's training texts, any text even vaguely
similar to Dante's production would be recognised as Dantean, the
classifier being untrained to make subtle distinctions. Instead, one
can expect better results if the classifier is trained to spot minimal
differences. We have thus done a large-scale screening of authors who
have written in Latin around the same historical period of Dante's,
\wasblue{and who have written works of either an epistolary or
literary nature};
since the included authors are close to each other, in the
above-mentioned cultural-stylistic sense, the two resulting datasets
are challenging ones for computational authorship analysis systems.

While we used \medone\ and \medtwo\ as training sets for our Ep13
work, of course they can be used as datasets for medieval Latin AV
research that does not necessarily involve Ep13 \wasblue{(we will
discuss such an example in Section~\ref{sec:twocasestudies})}, or as
datasets for other authorship analysis tasks that address medieval
Latin, \wasblue{or as benchmarks for general-purpose,
language-agnostic authorship verification systems}. This is the reason
why we make them available to the research community.
 

\subsection{Composition and preprocessing of the datasets}
\label{sec:composition}

\noindent The composition of our two datasets is described in detail
in Tables~\ref{tab:medone} and~\ref{tab:medtwo}.

\medone\ is composed of texts of epistolary genre (given that this is
the nature of Ep13(I)) mostly dating back to the 13th and 14th
centuries, for a total of 294 epistles; the average length of these
epistles is 378 words. Most of the texts are actually entire
collections of epistles; we consider each epistle as a single training
text. Note that, concerning the epistles by Guido Faba and Pietro
della Vigna (rows 4 and 5 of Table~\ref{tab:medone}), we have not used
the entire collections available from
\citeprim{Dangelo:2014jn,Gaudenzi:1971hb}, but only parts of them. One
reason is that some such epistles are extremely short in length
(sometimes even a single sentence), and hence they would not have
conveyed much information to the training process. The second reason
is that, as can be seen in Table~\ref{tab:medone}, Guido Faba and
Pietro della Vigna are the two authors for whom we have the highest
number of epistles anyway, and including the collections in their
entirety would have made the dataset even more imbalanced than it
already is.

\medtwo\ contains instead (given the similar nature of Ep13(II)) texts
of a non-epistolary nature, especially exegetic comments on literary
works and treatises, also dating to the 13th and 14th centuries, for a
total of 30 texts; the average length of these texts is 39,958 words,
i.e., about 100 times longer (on average) than those of \medone. Some
of these texts are not included in their entirety. In these cases, the
portions excluded mainly consist of lengthy \emph{explicit} citations
to other authors' works; as already mentioned in
Section~\ref{sec:relatedwork}, we have removed explicit citations
since they provide noise, rather than information, to an authorship
analyser.

All of the texts included in the two datasets are such that their
authorship is certain, i.e., is not currently disputed by any
scholar.\footnote{Note that from Petrus de Boateriis' collection (see
last row of Table~\ref{tab:medone}) we have removed the epistle
allegedly written by Cangrande della Scala to Henry VII, since it has
recently been suggested (see Footnote~\ref{foot:pellegrini}
\wasblue{and Section~\ref{sec:twocasestudies}}) that it may have been
written by Dante Alighieri.} Some of the texts were already available
in .txt format, and their inclusion in the dataset has thus posed no
major problem. Some other texts were only available in .pdf format, or
only on paper; in these cases, we converted the .pdf or the scanned
images into .txt format via an optical character recognition
software\footnote{FreeOCR, available at
\url{ http://www.paperfile.net/}~.}, and thoroughly corrected the
output by hand.

We have subjected all texts to a number of preprocessing steps
necessary for performing accurate authorship analysis; these include
\begin{itemize}
\item Removing any meta-textual information that has been inserted by
  the curator of the edition, such as titles, page numbers, quotation
  marks, square brackets, etc; this cleans the documents from obvious
  editorial intervention.
\item Marking explicit citations in Latin with asterisks, and explicit
  citations in languages other than Latin (mostly Florentine
  vernacular) with curly brackets; this is both to allow ignoring them
  in the computation (since they are the production of someone
  different than the author of the text) or to use them as a potential
  authorial-related feature (i.e., the usage of citations in different
  languages), at the discretion of the researcher.
\item Replacing every occurrence of the character ``v'' with the
  character ``u''; the reason for this lies in the different
  approaches followed by the various editors of the texts included,
  regarding whether to consider ``u'' and ``v'' as the same character
  or not.\footnote{In medieval written Latin there was only one
  grapheme, represented as a lowercase ``u'' and a capital ``V'',
  instead of the two modern graphemes ``u-U'' and ``v-V''.}
\end{itemize}

\noindent 
The two datasets are available for download at
\url{https://doi.org/10.5281/zenodo.4298503};
a \texttt{readme} file is also included that explains the structure of
the archive.\footnote{\label{foot:zenodo}Zenodo is an open-access
repository that provides free and permanent access to the resources
stored on it; see \url{https://about.zenodo.org/}.}

\begin{table}[t]
  \caption{Composition of the \medone\ dataset; the 3rd column
  indicates the approximate historical period in which the texts were
  written, the 4th and 5th columns indicate the number of texts and
  the number of words that the collection consists of, while
  \wasblue{the 7th and 8th columns indicate the $F_{1}$ value and the
  $Acc$ (``vanilla accuracy'') value} obtained in the experiments of
  Section~\ref{sec:results} by the authorship verifier trained via
  logistic regression for the specified author.}
  \begin{footnotesize}
    \begin{center}
      \resizebox{\textwidth}{!} {
      \begin{tabular}{|l|L|c|r|r|c|c|c|}
        \hline
        \multicolumn{1}{|c|}{\multirow{2}{*}{Author}} & \multicolumn{1}{c|}{\multirow{2}{*}{Text (or collection thereof)}} & Period & \multicolumn{1}{c|}{\multirow{2}{*}{\#d}} & \multicolumn{1}{c|}{\multirow{2}{*}{\#w}} & \multicolumn{1}{c|}{\multirow{2}{*}{Ed.}} & \multicolumn{1}{c|}{\multirow{2}{*}{$F_{1}$}} &                    \multicolumn{1}{c|}{\multirow{2}{*}{$Acc$}} \\
                                                      & & (approx.) & & & & & \\
        \hline \hline
        \multirow{2}{*}{Clara Assisiensis} & \textit{Epistola ad Ermentrudem} & 1240-1253 & 1 & 249 &  \citeprim{Menesto:1995hz} & \multirow{2}{*}{\wasblue{1.000}} & \multirow{2}{*}{\wasblue{1.000}} \\
        \cline{2-6} 
                                                      & \textit{Epistolae ad sanctam Agnetem de Praga} I, II, III & 1234-1253 & 3 & 1,842 & \citeprim{Menesto:1995hz} & & \\
        \hline
        Dante Alighieri & Epistles & 1304-1315 & 12 & 6,061 & \citeprim{Frugoni:1996qm} & \wasblue{0.857} & \wasblue{0.990}\\
        \hline
        Giovanni Boccaccio & Epistles and letters & 1340-1375 & 24 & 25,789 & \citeprim{Auzzas:1992df} & \wasblue{0.980} & \wasblue{0.997} \\ 
        \hline
        Guido Faba & Epistles& 1239-1241 & 78 & 7,203 & \citeprim{Gaudenzi:1971hb} & \wasblue{0.946} & \wasblue{0.973}\\
        \hline
        Pietro della Vigna & The collected epistles of Pietro della Vigna & 1220-1249 & 146 & 65,004 & \citeprim{Dangelo:2014jn} & \wasblue{0.986} & \wasblue{0.986}\\ \hline
        (Various authors) & Epistles from the collection of Petrus de Boateriis & 1250-1315 & 30 & 5,056 & \citeprim{Schneider:1926jw} & --- & --- \\ \hline
      \end{tabular}
      }
    \end{center}
  \end{footnotesize}
  \label{tab:medone}
\end{table}


\begin{table}[t]
 \caption{Composition of the \medtwo\ dataset; the meanings of the columns are as in Table~\ref{tab:medone}. 
 }
 \begin{scriptsize}
   \begin{center}
     \resizebox{\textwidth}{!} {
     \begin{tabular}{|l|L|c|r|c|c|c|}
       \hline
       \multicolumn{1}{|c|}{\multirow{1}{*}{Author}} & \multicolumn{1}{c|}{\multirow{1}{*}{Text}} & Period & \multicolumn{1}{c|}{\multirow{1}{*}{\#w}} & \multicolumn{1}{c|}{\multirow{1}{*}{Ed.}} & \multicolumn{1}{c|}{\multirow{1}{*}{$F_{1}$}} & \multicolumn{1}{c|}{\multirow{1}{*}{$Acc$}}\\
       \hline \hline
       Bene Florentinus & \textit{Candelabrum} & 1238 & 41,078 & \citeprim{Alessio:1983gh} & --- & ---\\ \hline
       \multirow{3}{*}{Benvenuto da Imola} & \emph{Comentum super Dantis Aldigherij Comoediam} & 1375-1380 & 105,096 & \citeprim{Brownlee:2018bm} & \multirow{3}{*}{\wasblue{0.800}} & \multirow{3}{*}{\wasblue{0.967}}\\ \cline{2-5}
                                                     & \textit{Expositio super Valerio Maximo} & 1380 & 3,419 & \citeprim{Rossi:2002ip} & & \\ \cline{2-5}
                                                     & \textit{Glose Bucolicorum Virgilii} & 1380 & 3,912 & \citeprim{Mancuso:2015tz} & & \\ \hline
       \multirow{4}{*}{Boncompagno da Signa} & \emph{Liber de obsidione Ancone} & 1198-1200 & 7,821 & \citeprim{Garbini:1999mq} & \multirow{4}{*}{\wasblue{0.333}} & \multirow{4}{*}{\wasblue{0.867}}\\ \cline{2-5}
                                                     & \textit{Palma} & 1198 & 5,022 & \citeprim{Sutter:1894rv} & & \\ \cline{2-5}
                                                     & \textit{Rota Veneris} & ante 1215 & 4,632 & \citeprim{Garbini:1996tf} & & \\ \cline{2-5}
                                                     & \textit{Ysagoge} & 1204 & 8,550 & \citeprim{Clark:1997yg} & & \\ \hline
       \multirow{2}{*}{Dante Alighieri} & \emph{De Vulgari Eloquentia} & 1304--1306 & 11,384 & \citeprim{Tavoni:2011by} & \multirow{2}{*}{\wasblue{0.500}} & \multirow{2}{*}{\wasblue{0.933}}\\ \cline{2-5}
                                                     & \textit{Monarchia} & 1313--1319 & 19,162 & \citeprim{Nardi:1996fd} & & \\ \hline
       Filippo Villani & \textit{Expositio seu comentum super Comedia Dantis Allegherii} & 1391-1405 & 31,503 & \citeprim{Ferrario:1999nf} & --- & --- \\
       \hline
       \multirow{3}{*}{Giovanni Boccaccio} & \emph{De vita et moribus d. Francisci Petracchi} & 1342 & 1,884 
                                                                                                                                                       & \citeprim{Fabbri:1992hv} & \multirow{3}{*}{\wasblue{0.800}} & \multirow{3}{*}{\wasblue{0.967}}\\ \cline{2-5}
                                                     & \textit{De mulieribus claris} & 1361-1362 & 49,242 & \citeprim{Zaccaria:1967oa}
                                                                                                                                                                                                   & & \\ \cline{2-5}
                                                     & \textit{De Genealogia deorum gentilium} & 1360-1375 & 198,508 & \citeprim{Romano:1951uz} & & \\ \hline
       \multirow{2}{*}{Giovanni del Virgilio} & \textit{Allegorie super fabulas Ovidii Methamorphoseos} & 1320 & 25,131 & \citeprim{Cotza:2013ae} & \multirow{2}{*}{\wasblue{0.000}} & \multirow{2}{*}{\wasblue{0.933}} \\ \cline{2-5}
                                                     & \emph{Ars dictaminis} & 1320 & 2,376 & \citeprim{Kristeller:1961la} & & \\ \hline
       Graziolo Bambaglioli & A Commentary on Dante's Inferno & 1324 & 41,104 & \citeprim{Rossi:1999hq} & --- & --- \\ \hline
       Guido da Pisa & \textit{Expositiones et glose. Declaratio super Comediam Dantis} & 1327-1328 & 87,822 & \citeprim{Cioffari:1974bg} & --- & --- \\ \hline
       Guido de Columnis & \textit{Historia destructionis Troiae} & 1272-1287 & 82,753 & \citeprim{Griffin:1936ka} & --- & --- \\ \hline
       Guido Faba & \textit{Dictamina rhetorica} & 1226-1228 & 16,982 & \citeprim{Gaudenzi:1971qu} & --- & --- \\ \hline
       Iacobus de Varagine & \textit{Chronica civitatis Ianuensis} & 1295-1298 & 53,864 & \citeprim{Monleone:1941ua} & --- & --- \\ \hline
       Iohannes de Appia & \textit{Constitutiones Romandiolae} & 1283 & 4,068 & \citeprim{Baldeschi:1926dn} & --- & --- \\ \hline
       Iohannes de Plano Carpini & \textit{Historia Mongalorum} & 1247-1252 & 20,145 & \citeprim{Daffina:1989lm} & --- & --- \\ \hline
       Iulianus de Spira & \textit{Vita Sancti Francisci} & 1232-1239 & 12,396 & \citeprim{Menesto:1995hz} & --- & --- \\ \hline
       \multirow{2}{*}{Nicola Trevet} & \emph{Expositio Herculis Furentis} & 1315-1316 & 33,017 & \citeprim{Ussani:1959tg} & \multirow{2}{*}{\wasblue{1.000}} & \multirow{2}{*}{\wasblue{1.000}} \\ \cline{2-5}
                                                     & \textit{Expositio L. Annaei Senecae Agamemnonis} & 1315-1316 & 19,873 & \citeprim{Meloni:1953rb} & & \\ \hline
       Pietro Alighieri & \textit{Comentum super poema Comedie Dantis} & 1340-1364 & 186,608 & \citeprim{Chiamenti:1999sw} & --- & --- \\ \hline
       Ryccardus de Sancto Germano & \textit{Chronicon} & 1216-1243 & 36,525 & \citeprim{Garufi:1937ja} & --- & --- \\ \hline
       Raimundus Lullus & \textit{Ars amativa boni} & 1290 & 82,733 & \citeprim{Romano:2004ap} & --- & --- \\ \hline
       Zono de’ Magnalis & Life of Virgilio & 1340 & 2,136 & \citeprim{Stok:1991ft} & --- & --- \\ \hline
     \end{tabular}
     }
   \end{center}
 \end{scriptsize}
 \label{tab:medtwo}
\end{table}%


\section{Baseline Authorship Verification Results}
\label{sec:results}

\noindent In \citesec{Corbara:2019cq} we briefly describe some
authorship verification experiments that we have run on \medone\ and
\medtwo. \wasblue{For the present paper we have rerun the experiments
completely, revising and correcting the experimental protocol that we
had followed in \citesec{Corbara:2019cq}.\footnote{\label{foot:loo}In
\citesec{Corbara:2019cq} we had performed both feature selection and
parameter optimisation on the entire dataset, and we had subsequently
estimated the accuracy of the system by applying the leave-one-out
protocol. This means that, when a document was used as the test
document, it had already participated both in the feature selection
process and in the parameter optimisation process, which are parts of
the training process; this is not legitimate.  Thanks to two anonymous
reviewers for pointing this out.} As a consequence, there are slight
differences between the accuracy values reported in
\citesec{Corbara:2019cq} and those reported here.} In order to ease
the task of researchers wishing to replicate and/or to outperform the
results we have obtained,
we make available at \url{https://doi.org/10.5281/zenodo.3903235}
the source code of \medievalla, the authorship verification tool that
we have developed and used in order to obtain these results.

For these experiments, first of all we remove explicit citations,
either in Latin or other languages, and we segment each resulting text
into shorter texts, so as to increase the overall number of labelled
texts, while reducing their average size. This is necessary because
machine learning processes require a significant number of training
examples, regardless of their length. In particular, for each text:
\begin{itemize}

\item we identify the sentences that make up the text (using the NLTK
  package, available at \url{https://www.nltk.org/}); if a sentence is
  shorter than 8 words, we merge it with the next sentence (or the
  previous sentence, if it is the last sentence of the text);

\item we create sequences of 3 consecutive sentences (hereafter:
  ``segments''), consider each of these sequences as a labelled text,
  and assign it the author label of the text from which it was
  extracted.

\end{itemize}
\noindent Following this process, we use as labelled texts both the
original texts in their entirety \emph{and} the segments.  Thus, the
number of labelled texts has increased from 294 to 1,310 for \medone\
and from 30 to 12,772 for \medtwo.

For our experiments, we lower-case the entire text, remove punctuation
marks, and convert each labelled text into a vector of features. The
reason why we ignore punctuation marks is that they were not inserted
by the authors (punctuation was absent or hardly coherent in ancient
manuscripts, and such marks have been introduced into texts by
editors).

The set of features we use is subdivided into six subsets of different
feature types:

\begin{enumerate}
\item Character $n$-grams ($n\in\{3,4,5\}$);
\item Word $n$-grams ($n\in\{1,2\}$);
\item Function words (from a list of 74 Latin function words);
\item Verbal endings (from a list of 245 regular Latin verbal
  endings);
\item Word lengths (from 1 to 23 characters);
\item Sentence lengths (from 3 to 70 words).
\end{enumerate}
\noindent and the vector space results from the union of all of these
features. In order to deal with the high dimensionality of the feature
space we subject the features resulting in a sparse distribution
(character $n$-grams and word $n$-grams) to a process of
dimensionality reduction. First, we perform feature selection via the
Chi-square function 
\wasblue{(see
e.g.,~\citesec{Yang97})},
where probabilities are interpreted on the event space of documents;
in other words, $\Pr(t_{k},a_{j})$ represents the probability that,
for a random document that belongs to class $a_{j}$ (i.e., that was
written by author $a_{j}$), feature $t_{k}$ appears in the
document. In our experiments we select the best 10\% character
$n$-grams and the best 10\% word $n$-grams.
%
We then perform feature weighting via the
$\tfidf$ function in its standard ``ltc'' variant \wasblue{(see
e.g.,~\citesec{Salton88})}.
%
For \medone\ the number of resulting features is 16,101, while for
\medtwo\ this number is instead 86,924.

\wasblue{
The six subsets of features
described above have very different cardinality: the numbers of
features contained in sets (1) and (2) depend on the dataset, but is
in general very high (in both cases it typically ranges in the tens --
or hundred -- thousands features), while the numbers of features
contained in sets (3), (4), (5), (6) are fixed (there are 74, 245, 23,
68, features in each of these groups, respectively), and are much
smaller than the two previous ones. This means that the latter groups
may end up being overwhelmed, in terms of their contribution to the
verification process, by the former groups. In order to avoid this,
we individually normalise each of the six feature subsets via
L2-normalisation, so that each of the six vectors subspaces they
define have unit norm.\footnote{\wasblue{This means that the
contribution of, say, a character $n$-gram, ends up being
smaller than the contribution of, say, a word length, because there
are more character $n$-grams than word lengths.
This does not prevent the classifier from uncovering which among the
features are the most important (these might well include some
character $n$-grams) or least important (these might well include some
word lengths), though, since the classifier attempts to find the
linear combination of feature weights that best classifies the
documents}.}}

As the learning mechanism we use \textit{logistic regression}, as
implemented in the \texttt{scikit-learn}
package.\footnote{\url{https://scikit-learn.org/stable/index.html}} We
train each binary classifier by optimising hyperparameter $C$ (the
inverse of the regularisation strength) via stratified 10-fold
cross-validation (10-FCV), using a grid search on the set \{0.001,
0.01, \ldots, 100, 1000\}. \wasblue{We use a variant of stratified
10-FCV called ``grouped'' stratified 10-FCV, that prevents different
segments from the same document (``group'') to end up in different
folds; in this way, the classifier never unduly benefits from testing
on segments of a document, other segments of which have been seen
during training.}
\wasblue{There are two main reasons why we have used logistic
regression. One is the fact that it generates classifiers that have
proven very effective across a broad spectrum of text classification
scenarios. A second reason is the fact that, together with a binary
classification decision, for each document $d$ it returns a
``confidence score'' (i.e., a measure of the confidence that the
classifier has in the correctness of its own decision) in the form of
a probability value (called a ``posterior probability''), and that
these probability values tend to be \emph{well calibrated} (i.e.,
reliable probability values)~\citesec{Niculescu-Mizil:2005gh}.}

\wasblue{We also briefly report on some additional experiments for
which we have used other learning algorithms, i.e., SVMs (for which we
have optimised hyperparameter $C$ via grid search on \{0.001, 0.01,
\ldots, 100, 1000\}) and multinomial naive Bayes (for which we have
optimised parameter $\alpha=\{10^i\}$ for
$i\in \{-7,-6,\ldots,-1, 0\}$ ).}

We subject the resulting \medievalla\ system to a ``leave-one-out''
validation test, which consists of predicting, for each dataset
$D\in\{\medone,\medtwo\}$, for each author $a$ in the set of authors
$\mathcal{A}$ represented in $D$, and for each document $d\in D$,
whether $a$ is the author of $d$ or not, where the prediction is
issued by an ``$a$ vs.\ (NOT $a$)'' binary classifier trained on all
labelled texts (i.e., segments \emph{and} entire documents) from
\wasblue{$D\setminus\{d\}$}. This means that all labelled texts from
documents in \wasblue{$D\setminus\{d\}$} originating from author $a$
are used as positive training examples while all labelled texts from
documents in \wasblue{$D\setminus\{d\}$} originating from authors
other than $a$ are used as negative training examples. Note that
\begin{itemize}

%
\item In order to faithfully reproduce the operating conditions of an
  authorship verifier, as test examples we use only entire documents,
  i.e., we use segments and entire documents for training purposes but
  only entire documents for testing purposes.

\item In order to avoid any overlap between training examples and test
  examples, when document $d$ is used as a test document we exclude
  from the training set all the segments derived from $d$.
  
\item \wasblue{In order to avoid any overlap between the training
  phase and the test phase, both the feature selection step and the
  parameter optimisation step are performed not on the entire dataset
  $D$, but on $D\setminus\{d\}$. This means that the entire cycle
  (feature selection + parameter optimisation + classifier training)
  is repeated for each document $d\in D$, for both \medone\ and
  \medtwo.}

\item We have not generated classifiers for authors for which we have
  only one text in $D$, since this would entail experiments in which
  the author is not present both in the training and in the test
  set;\footnote{\wasblue{For the very same reason, we bypass the
  parameter optimisation phase in cases in which we only have 2
  positive documents and one of them is acting as the held-out
  document. This causes the training set to have only one positive
  document (plus fragments) and this eventually forces one of the
  trainings (as generated via 10-fold cross-validation) to be devoid
  of any positive example (since in the ``grouped'' variant of
  stratified 10-FCV the fragments of a document are always within the
  same fold as the full document, for reasons already discussed). In
  those (few) cases, we resort to a logistic regressor that is
  moderately regularised (we set $C=0.1$) in order to avoid
  overfitting the one and only positive document; likewise, for SVMs
  we also set $C=0.1$ and for multinomial naive Bayes we set
  $\alpha=0.001$.}}  as a result, the texts of these authors are used
  only as negative examples in experiments centred on other
  authors. Ultimately, this means that we have trained binary
  classifiers for 5 authors of \medone\ (all authors except those from
  the collection of Petrus de Boateriis, since this collection is a
  miscellanea of authors) and 6 authors for \medtwo; this leads to
  5$\times$294=1470 predictions for \medone\ and 6$\times$30=180
  predictions for \medtwo, \wasblue{where each prediction is the
  result of a different cycle consisting of feature selection +
  parameter optimisation + classifier
  training.\footnote{\wasblue{\label{foot:trainings}Since we use
  10-fold cross validation for parameter optimisation and explore a
  grid of 7 parameters, or experimentation consists of roughly 115,000
  trainings per learner (we report experiments for 3 learners).}}}

\end{itemize}
\noindent In order to evaluate the performance of a binary AV system
we use, as customary, the $F_{1}$ function, defined as
\begin{align}
  \label{eq:F1}
  F_{1} = & \ \left\{
            \begin{array}{cl}
              \dfrac{2\TP}{2\TP + \FP + \FN} & \mathrm{if} \ \TP + \FP + \FN>0 \rule[-3ex]{0mm}{7ex} \\
              1 & \mathrm{if} \ \TP=\FP=\FN=0 \\
            \end{array}
  \right.
\end{align}
\noindent where $\TP$, $\FP$, $\FN$, represent the numbers of true
positives, false positives, false negatives, generated by the binary
AV system. $F_{1}$ ranges between 0 (worst) and 1 (best). In order to
compute $F_{1}$ across an entire dataset, for which several binary AV
systems need to be deployed (5 for \medone\ and 6 for \medtwo), we
compute its \emph{macroaveraged} variant (denoted by $F_{1}^{M}$) and
its \textit{microaveraged} variant (denoted by
$F_{1}^{\mu}$). $F_{1}^{M}$ is obtained by first computing values of
$F_{1}$ for all $a_{j}\in\mathcal{A}$ and then averaging
them. $F_{1}^{\mu}$ is obtained by (a) computing the author-specific
values $\TP_{j}$, $\FP_{j}$, $\FN_{j}$ for all $a_{j}\in\mathcal{A}$;
(b) obtaining $\TP$ as the sum of the $\TP_{j}$'s (same for $\FP$ and
$\FN$), and then (c) applying Equation~\ref{eq:F1}.
\wasblue{For completeness we also report effectiveness results in
terms of the so-called ``vanilla accuracy'' measure, defined as
\begin{align}
  \label{eq:acc}
  Acc = & \ \dfrac{\TP+\TN}{\TP+\FP+\FN+\TN}
\end{align}

\noindent i.e., as the ratio between the number of correct predictions
and the number of predictions. In order to compute $Acc$ across
different binary AV systems, either the microaveraged or the
macroaveraged version of $Acc$ can be computed, along the same lines
as for $F_{1}$. Unlike for $F_{1}$, though, the microaveraged and the
macroaveraged versions of $Acc$ are demonstrably the same measure,
which we will thus simply indicate as $Acc$, without $\mu$ or $M$
superscripts.  }

Our experimental results are reported in
Table~\ref{tab:results}.\footnote{\label{foot:tworeasons}\wasblue{Two
further reasons why these} results slightly differ from the ones
reported in \citesec{Corbara:2019cq} \wasblue{are that} (a) some
\texttt{scikit-learn} libraries that we use are now available in
updated versions, different from the ones we had used
in~\citesec{Corbara:2019cq}; (b) the stratified 10-fold
cross-validation that we use for optimizing hyperparameter $C$ splits
the data into 10 folds randomly, and this random component can
introduce small fluctuations in the final results. Overall, these
fluctuations are noticeable but not substantial from a qualitative
point of view. The results we report in this paper should be exactly
reproducible (barring changes in \texttt{scikit-learn} libraries) by
anyone who downloads the code and the datasets, also thanks to the
fact that we have now ``seeded'' the stratified 10-fold
cross-validation process, thus eliminating the above-mentioned random
component.}
\begin{table}[t]
  \caption{Summary results of our AV experiments on the \medone\ and
  \medtwo\ datasets.}
  \begin{center}
    \begin{tabular}{c||c|c|c||c|c|c}
      & \multicolumn{3}{c||}{\medone} & \multicolumn{3}{c}{\medtwo} \\
      \hline
      \rule[-2ex]{0mm}{5ex} 
      Learner & $F_{1}^{M}$ & $F_{1}^{\mu}$ & \wasblue{$Acc$} & $F_{1}^{M}$ & $F_{1}^{\mu}$ & \wasblue{$Acc$} \\
      \hline    \hline
      LR & \wasblue{0.954} & \wasblue{0.969} & \wasblue{0.989} & \wasblue{0.572} & \wasblue{0.615} & \wasblue{0.944} \\
      SVM & \wasblue{0.944} & \wasblue{0.969} & \wasblue{0.989} & \wasblue{0.383} & \wasblue{0.435} & \wasblue{0.928} \\
      MNB & \wasblue{0.760} & \wasblue{0.933} & \wasblue{0.976} & \wasblue{0.310} & \wasblue{0.357} & \wasblue{0.900} \\
    \end{tabular}
  \end{center}
  \label{tab:results}
\end{table}%
The last columns of Tables~\ref{tab:medone} and \ref{tab:medtwo}
report the $F_{1}$ and $Acc$ values we have obtained for the
individual authors for which we have generated binary AV systems; from
these it is easy to compute the $F_{1}^{M}$ values and average $Acc$
values of Table~\ref{tab:results} \wasblue{by simply averaging them}.

\wasblue{
Note that, as evident from the
$F_{1}$ and $Acc$ columns of Tables~\ref{tab:medone} and
\ref{tab:medtwo}, there is a lot of variability in the scores
(especially for $F_{1}$) across different authors for the same
dataset. There are at least three possible explanations for this:}
  
\wasblue{
\begin{itemize}
  
\item For some authors there are more (positive) training data than
  for other authors. Since authorship verification consists of a
  different binary classification task for each author, this means
  that it will be easier (other things being equal) to conduct
  authorship verification for the former authors than for the latter.
    
\item Some large differences in $F_{1}$ values are due to the
  idiosyncrasies of the $F_{1}$ measure. For instance, the authorship
  verifier for Giovanni del Virgilio (see Table~\ref{tab:medtwo}),
  when asked to verify the 30 texts in \medtwo, returns 2 false
  negatives and 28 true negatives. Despite having correctly predicted
  28 out of 30 times (the ``vanilla accuracy'' result is
  $Acc$=28/30=0.933), the verifier obtains an $F_{1}$ value of 0
  because (see Equation~\ref{eq:F1}) there are no true positives,
  i.e., none of the two texts actually by Giovanni del Virgilio were
  correctly predicted as by him.
  
\item Even if we had the same quantity of training data for each
  author,
  we might obtain different accuracy results for different authors
  because some authors may inherently be more difficult to identify,
  from a stylistic point of view, than others.
\end{itemize}
}

\noindent At 
\wasblue{\url{https://doi.org/10.5281/zenodo.4298503}}
we provide, in spreadsheet form, the list of all $\langle$author,
document$\rangle$ \blue{classification decisions as taken} by
\medievalla, as well as the $F_{1}^{\mu}$ results that are also
reported in Table~\ref{sec:results} and the author-specific $F_{1}$
values also reported in Tables~\ref{tab:medone} and \ref{tab:medtwo}.

\blue{Interestingly enough, an analysis of these individual
classification decisions shows that \emph{there are no systematic
mistakes}, but just a few, scattered individual ones. More in
particular, it never happens that there are two or more incorrectly
classified documents with the same true author $A_{1}$ \emph{and} with
the same predicted author $A_{2}$, with $A_{1} \not = A_{2}$; in other
words, there are no systematic mistakes that would indicate an extreme
similarity in style between two authors $A_{1}$ and $A_{2}$. One of
the reasons for this is that the mistakes made by our verifiers are
very few, i.e., only 26 out of 1650 verification decisions (16 out of
1470 for the \medone\ experiments and 10 out of 180 for the \medtwo\
experiments) are incorrect.}


\wasblue{
\section{Two disputed epistles}
\label{sec:twocasestudies}


\subsection{The Epistle to Cangrande}
\label{sec:Cangrande}

\noindent In Section~\ref{sec:origin} we mentioned that the original
reason for developing these two datasets was the attempt to solve the
puzzle of the \emph{Epistle to Cangrande}, i.e., verifying if the
letter addressed to Cangrande della Scala was indeed written by Dante
Alighieri. After running the experiments described in
Section~\ref{sec:results}, for each of the two datasets we have
retrained the authorship verifier for author Dante Alighieri (i.e.,
the one that whose \textsc{Yes} label indicates authorship by Dante
and whose \textsc{No} label indicates authorship by someone other than
Dante), rerunning the entire cycle ``feature selection + parameter
optimisation + classifier training'' on the entire dataset; we have
then applied the classifier derived from \medone\ to the first portion
of the epistle (Ep13(I)) and the classifier derived from \medtwo\ to
the second portion (Ep13(II)).

\begin{table}[t]
  \caption{Results of the application of the two authorship verifiers
  ``Dante vs.\ Not Dante'' to the two portions of the Epistle to
  Cangrande. Columns 4 and 5 recall (from Tables~\ref{tab:medone}
  and~\ref{tab:medtwo}) the $F_{1}$ \wasblue{and $Acc$} values that
  the ``Dante vs.\ Not Dante'' verifiers have obtained in the
  experiments of Section~\ref{sec:results}.}
  \begin{center}
    \begin{tabular}{c||c|c|c|c}
      & \textsf{Binary decision} & \textsf{Posterior probability} & $F_{1}$ &  $\wasblue{Acc}$ \\
      \hline
      Ep13(I)  & No & \wasblue{0.367} & \wasblue{0.857} & \wasblue{0.990} \\
      Ep13(II) & No & \wasblue{0.022} & \wasblue{0.500} & \wasblue{0.933} \\
    \end{tabular}
  \end{center}
  \label{tab:part1results}
\end{table}

The results of the application of the two classifiers are reported in
Table~\ref{tab:part1results}.  These results show that out authorship
verifiers believe that both portions of the Epistle to Cangrande are
the work of a malicious forger.}

\wasblue{Once applied to Ep13(I), the ``Dante vs.\ Not Dante''
verifier trained on \medone\ returns a posterior probability of 0.367:
this means that the verifier believes that Ep13(I) is not by Dante
(since this probability is <0.500), and is moderately confident about
this fact (its ``degree of confidence'' being (1-0.367)=0.633).  As
from Table~\ref{tab:medone}, this verifier has also proved very
accurate ($F_{1}=0.857$, $Acc=0.990$) once tested on \medone\ via
leave-one-out. These two facts, altogether, make a fairly convincing
case for the non-Dantean authorship of Ep13(I).

Concerning Ep13(II), instead, once applied to it, the ``Dante vs.\ Not
Dante'' verifier trained on \medtwo\ returns a posterior probability
of 0.022: this means that the verifier believes that Ep13(II) is also
not by Dante (since this probability is <0.500), and is extremely
confident about this fact (its degree of confidence being
(1-0.022)=0.978).  As from Table~\ref{tab:medtwo}, this verifier has
proved reasonably accurate ($F_{1}=0.500$, $Acc=0.933$) once tested on
\medtwo\ via leave-one-out. These two facts support the hypothesis
that also Ep13(I) is not by Dante.\footnote{\wasblue{Note that a
classifier that obtains $F_{1}=0.500$ is \emph{not} equivalent to a
classifier that returns random decisions: in fact, a completely
clueless classifier for which half of the positives are true positives
while the other half are false negatives, and half of the negatives
are true negatives while the other half are false positives, on
\medtwo\ would obtain a value of
$F_{1}=(2\TP)/(2\TP+\FP+\FN)=(2\cdot 1)/(2\cdot 1 + 14 + 1)=0.058$.
The $F_{1}=0.500$ result for the ``Dante vs.\ Not Dante'' authorship
verifier is the result of generating, on dataset \medtwo, 1 true
positive, 1 false positive, 1 false negative, and 27 true negatives,
i.e., 28 correct predictions out of 30 total predictions.}}}



\wasblue{
\subsection{The Epistle to Henry VII}
\label{sec:HenryVII}

\noindent While we were carrying out our research on Ep13 that led to
the creation of \medone\ and \medtwo, a paper
appeared~\citesec{Pellegrini:2018yb} whose object was an epistle
addressed to emperor Henry VII and signed by Cangrande della
Scala. The author of~\citesec{Pellegrini:2018yb}, based on an analysis
of the contents of the epistle, conjectured that its author could be
Dante Alighieri himself. Since we had already trained a ``Dante vs.\
not Dante'' authorship verifier on \medone, and since the texts
contained in \medone\ have also an epistular nature, it seemed natural
to preprocess the epistle to Henry VII in the same way as described in
Section~\ref{sec:results}, and apply to it the verifier trained on
\medone. The results of the application are described in
Table~\ref{tab:henryVIIresult}.

Our authorship verifier rejects the hypothesis that the epistle to
Henry VII may have been written by Dante, and is extremely confident
in its own prediction (i.e., it believes that the epistle is by
someone other than Dante with probability (1-0.026)=0.974). Together
with the fact that this verifier has shown very high accuracy
($F_{1}=0.857$, $Acc=0.990$) in the experiments of
Section~\ref{sec:results}, this makes us decidedly lean towards the
hypothesis that the epistle is not the work of Dante.

\begin{table}[t]
  \caption{\wasblue{Result of the application of the authorship
  verifier ``Dante vs.\ Not Dante'' to the Epistle to Henry
  VII. Column 4 recalls (from Table~\ref{tab:medone}) the $F_{1}$
  value that the ``Dante vs.\ Not Dante'' verifier has obtained in the
  experiments of Section~\ref{sec:results}.}}
  \begin{center}
    \begin{tabular}{c||c|c|c|c|}
      & \textsf{Binary decision} & \textsf{Posterior probability} & \textsf{$F_{1}$} & $Acc$ \\
      \hline
      EpHenryVII  & No & \wasblue{0.026} & 0.857 & 0.990 \\
    \end{tabular}
  \end{center}
  \label{tab:henryVIIresult}
\end{table}%

}


\section{Conclusion}
\label{sec:conclusion}

\noindent We have described \medone\ and \medtwo, two new datasets of
cultural heritage texts written in medieval Latin by 13th- and
14th-century (mostly Italian)
literates and labelled by author, that we make publicly available to
researchers working on computational authorship analysis. These
datasets can be valuable tools for researchers investigating
techniques for authorship attribution, authorship verification, or
same-authorship verification, especially for texts written in Latin or
medieval Latin.

We also make available the source code of \medievalla, an authorship
verification tool that we have built in order to work on an important
case study, i.e., the real paternity of the ``Epistle to Cangrande'',
allegedly written by Dante Alighieri but believed by some to be a
forgery.  We also describe in detail experiments (corrected versions
of the ones which we had reported in~\citesec{Corbara:2019cq}) in
which we have applied \medievalla\ to \medone\ and \medtwo. We hope
that the availability of the datasets (and of our authorship
verification tool) will allow researchers interested in authorship
verification to replicate our results, and possibly to outperform them
via improved AV techniques.


\section*{Acknowledgments}
\label{sec:acks}

\noindent We would like to thank Gabriella Albanese and Paolo Pontari
for helping us to identify the medieval Latin texts that we have
incorporated into our datasets;
Patrick Juola, Moshe Koppel, Vincenzo Mele, and Efstathios Stamatatos,
for suggesting important bibliographical references; and Carlo Meghini
for giving the initial impetus to this research and for stimulating
discussions on the topics covered by this article. \wasblue{Thanks to
the anonymous reviewers for useful feedback and for spotting two
problems in the experimental protocol used
in~\citesec{Corbara:2019cq}.}






\begin{thebibliography}{35}


\ifx \showCODEN    \undefined \def \showCODEN     #1{\unskip}     \fi
\ifx \showDOI      \undefined \def \showDOI       #1{#1}\fi
\ifx \showISBNx    \undefined \def \showISBNx     #1{\unskip}     \fi
\ifx \showISBNxiii \undefined \def \showISBNxiii  #1{\unskip}     \fi
\ifx \showISSN     \undefined \def \showISSN      #1{\unskip}     \fi
\ifx \showLCCN     \undefined \def \showLCCN      #1{\unskip}     \fi
\ifx \shownote     \undefined \def \shownote      #1{#1}          \fi
\ifx \showarticletitle \undefined \def \showarticletitle #1{#1}   \fi
\ifx \showURL      \undefined \def \showURL       {\relax}        \fi
\providecommand\bibfield[2]{#2}
\providecommand\bibinfo[2]{#2}
\providecommand\natexlab[1]{#1}
\providecommand\showeprint[2][]{arXiv:#2}

\bibitem[\protect\citeauthoryear{Alessio}{Alessio}{1983}]%
        {Alessio:1983gh}
\bibfield{author}{\bibinfo{person}{Gian~Carlo Alessio}.}
  \bibinfo{year}{1983}\natexlab{}.
\newblock \bibinfo{booktitle}{\emph{{Bene Florentini Candelabrum}}}.
\newblock \bibinfo{publisher}{Editrice Antenore}, \bibinfo{address}{Padova,
  IT}.
\newblock
\newblock
\shownote{{\url{https://bit.ly/2Xbl0pR} (Archivio della Latinità Italiana del
  Medioevo), accessed 2018-05-28}.}


\bibitem[\protect\citeauthoryear{Auzzas}{Auzzas}{1992}]%
        {Auzzas:1992df}
\bibfield{author}{\bibinfo{person}{Ginetta Auzzas}.}
  \bibinfo{year}{1992}\natexlab{}.
\newblock \bibinfo{booktitle}{\emph{Tutte le opere di {Giovanni Boccaccio}}}.
\newblock \bibinfo{publisher}{Mondadori}, \bibinfo{address}{Milano, IT},
  Chapter ``Epistole e lettere’’.
\newblock
\newblock
\shownote{{\url{https://bit.ly/2I2VLjp} (Biblioteca Italiana), accessed
  2018-05-28}.}


\bibitem[\protect\citeauthoryear{Baldeschi}{Baldeschi}{1926}]%
        {Baldeschi:1926dn}
\bibfield{author}{\bibinfo{person}{Luigi~Colini Baldeschi}.}
  \bibinfo{year}{1925-1926}\natexlab{}.
\newblock \showarticletitle{{Le ``Constitutiones Romandiolae" di Giovanni
  d'Appia}}.
\newblock \bibinfo{journal}{\emph{Nuovi Studi Medievali}} \bibinfo{volume}{2},
  \bibinfo{number}{1} (\bibinfo{year}{1925-1926}), \bibinfo{pages}{221--252}.
\newblock
\newblock
\shownote{{\url{https://bit.ly/30RRoAg} (Archivio della Latinità Italiana del
  Medioevo), accessed 2018-05-28}.}


\bibitem[\protect\citeauthoryear{Brownlee and Hollander}{Brownlee and
  Hollander}{2018}]%
        {Brownlee:2018bm}
\bibfield{author}{\bibinfo{person}{Kevin Brownlee} {and}
  \bibinfo{person}{Robert Hollander}.} \bibinfo{year}{2018}\natexlab{}.
\newblock \bibinfo{title}{{Benevenuti de Rambaldis de Imola Comentum super
  Dantis Aldigherij Comoediam, nunc primum integre in lucem editum sumptibus
  Guilielmi Warren Vernon, curante Jacobo Philippo Lacaita. Florentiae, G.
  Barbèra, 1887}}.
\newblock \bibinfo{howpublished}{{\url{https://bit.ly/2Dqj6du} (Darmouth Dante
  Project), accessed 2018-05-28}}.
\newblock


\bibitem[\protect\citeauthoryear{Chiamenti}{Chiamenti}{1999}]%
        {Chiamenti:1999sw}
\bibfield{author}{\bibinfo{person}{Massimiliano Chiamenti}.}
  \bibinfo{year}{1999}\natexlab{}.
\newblock \bibinfo{booktitle}{\emph{I commenti danteschi dei secoli {XIV, XV e
  XVI}}}.
\newblock \bibinfo{publisher}{LEXIS Progetti Editoriali},
  \bibinfo{address}{Roma, IT}, Chapter ``Comentum super Comedie Dantis (terza
  ed ultima redazione del `Comentum’)’’.
\newblock
\newblock
\shownote{{\url{https://bit.ly/2ECcU2c} (Biblioteca Italiana), accessed
  2018-05-28}.}


\bibitem[\protect\citeauthoryear{Cioffari}{Cioffari}{1974}]%
        {Cioffari:1974bg}
\bibfield{author}{\bibinfo{person}{Vincenzo Cioffari}.}
  \bibinfo{year}{1974}\natexlab{}.
\newblock \bibinfo{booktitle}{\emph{{Guido da Pisa's Expositiones et Glose
  super Comediam Dantis, or Commentary on Dante's Inferno}}}.
\newblock \bibinfo{publisher}{State University of New York Press},
  \bibinfo{address}{Albany, US}.
\newblock
\newblock
\shownote{{\url{https://bit.ly/2FRhT0x} (Darmouth Dante Project), accessed
  2018-05-28}.}


\bibitem[\protect\citeauthoryear{Clark}{Clark}{1997}]%
        {Clark:1997yg}
\bibfield{author}{\bibinfo{person}{Elmert Clark}.}
  \bibinfo{year}{1997}\natexlab{}.
\newblock \showarticletitle{{Magistri Boncompagni Ysagoge}}.
\newblock \bibinfo{journal}{\emph{Quadrivium}} \bibinfo{number}{8}
  (\bibinfo{year}{1997}), \bibinfo{pages}{23--71}.
\newblock
\newblock
\shownote{{\url{https://bit.ly/2MedBFc} (Archivio della Latinità Italiana del
  Medioevo), accessed 2018-05-28}.}


\bibitem[\protect\citeauthoryear{Cotza}{Cotza}{2013}]%
        {Cotza:2013ae}
\bibfield{author}{\bibinfo{person}{Valeria Cotza}.}
  \bibinfo{year}{2013}\natexlab{}.
\newblock \emph{\bibinfo{title}{{Giovanni del Virgilio, Allegorie super fabulas
  Ovidii Methamorphoseos. Edizione critica e introduzione}}}.
\newblock \bibinfo{thesistype}{Master's\ thesis}. \bibinfo{school}{Department
  of Philology, Literature and Linguistics, University of Pisa},
  \bibinfo{address}{Pisa, IT}.
\newblock


\bibitem[\protect\citeauthoryear{Daffinà, Leonardi, Lungarotti, Menestò, and
  Petech}{Daffinà et~al\mbox{.}}{1989}]%
        {Daffina:1989lm}
\bibfield{author}{\bibinfo{person}{Paolo Daffinà}, \bibinfo{person}{Claudio
  Leonardi}, \bibinfo{person}{Maria~Cristiana Lungarotti},
  \bibinfo{person}{Enrico Menestò}, {and} \bibinfo{person}{Luciano Petech}.}
  \bibinfo{year}{1989}\natexlab{}.
\newblock \bibinfo{booktitle}{\emph{{Giovanni di Pian di Carpine. Storia dei
  Mongoli}}}.
\newblock \bibinfo{publisher}{Fondazione CISAM}, \bibinfo{address}{Spoleto,
  IT}. 227--333 pages.
\newblock
\newblock
\shownote{{\url{https://bit.ly/2WoZaSM} (Archivio della Latinità Italiana del
  Medioevo), accessed 2018-05-28}.}


\bibitem[\protect\citeauthoryear{D'Angelo}{D'Angelo}{2014}]%
        {Dangelo:2014jn}
\bibfield{author}{\bibinfo{person}{Edoardo D'Angelo}.}
  \bibinfo{year}{2014}\natexlab{}.
\newblock \bibinfo{booktitle}{\emph{L'epistolario di {Pier della Vigna}}}.
\newblock \bibinfo{publisher}{Rubbettino Editore}, \bibinfo{address}{Soveria
  Mannelli, IT}.
\newblock


\bibitem[\protect\citeauthoryear{Fabbri}{Fabbri}{1992}]%
        {Fabbri:1992hv}
\bibfield{author}{\bibinfo{person}{Renata Fabbri}.}
  \bibinfo{year}{1992}\natexlab{}.
\newblock \bibinfo{booktitle}{\emph{Tutte le opere di {Giovanni Boccaccio}}}.
\newblock \bibinfo{publisher}{Mondadori}, \bibinfo{address}{Milano, IT},
  Chapter ``De vita et moribus d. Francisci Petracchi’’.
\newblock
\newblock
\shownote{{\url{https://bit.ly/2MdBUTV} (Biblioteca Italiana), accessed
  2018-05-28}.}


\bibitem[\protect\citeauthoryear{Ferrario}{Ferrario}{1999}]%
        {Ferrario:1999nf}
\bibfield{author}{\bibinfo{person}{Francesca Ferrario}.}
  \bibinfo{year}{1999}\natexlab{}.
\newblock \bibinfo{title}{Expositio seu comentum super {``Comedia” Dantis
  Allegherii}, a cura di Saverio Bellomo. Florence: Le Lettere, 1989.}
\newblock \bibinfo{howpublished}{{\url{https://bit.ly/2Uac1mS} (Darmouth Dante
  Project), accessed 2018-05-28}}.
\newblock


\bibitem[\protect\citeauthoryear{Frugoni and Brugnoli}{Frugoni and
  Brugnoli}{1996}]%
        {Frugoni:1996qm}
\bibfield{author}{\bibinfo{person}{Arsenio Frugoni} {and}
  \bibinfo{person}{Giorgio Brugnoli}.} \bibinfo{year}{1996}\natexlab{}.
\newblock \bibinfo{booktitle}{\emph{{Dante Alighieri - Opere minori}}}.
\newblock \bibinfo{publisher}{Riccardo Ricciardi Editore},
  \bibinfo{address}{Milano, IT}, Chapter ``Epistole’’.
\newblock
\newblock
\shownote{{\url{https://bit.ly/2JIPYTp} (Biblioteca Italiana), accessed
  2018-05-28}.}


\bibitem[\protect\citeauthoryear{Garbini}{Garbini}{1996}]%
        {Garbini:1996tf}
\bibfield{author}{\bibinfo{person}{Paolo Garbini}.}
  \bibinfo{year}{1996}\natexlab{}.
\newblock \bibinfo{booktitle}{\emph{{Boncompagnus de Signa - Rota Veneris}}}.
\newblock \bibinfo{publisher}{{Salerno Editrice}}, \bibinfo{address}{Roma, IT}.
\newblock
\newblock
\shownote{{\url{https://bit.ly/2wrCTVS} (Archivio della Latinità Italiana del
  Medioevo), accessed 2018-05-28}.}


\bibitem[\protect\citeauthoryear{Garbini}{Garbini}{1999}]%
        {Garbini:1999mq}
\bibfield{author}{\bibinfo{person}{Paolo Garbini}.}
  \bibinfo{year}{1999}\natexlab{}.
\newblock \bibinfo{booktitle}{\emph{{Boncompagnus de Signa - Liber de obsidione
  Ancone}}}.
\newblock \bibinfo{publisher}{Viella}, \bibinfo{address}{Roma, IT}.
\newblock
\newblock
\shownote{{\url{https://bit.ly/2QuMMLw} (Archivio della Latinità Italiana del
  Medioevo), accessed 2018-05-28}.}


\bibitem[\protect\citeauthoryear{Garufi}{Garufi}{1937}]%
        {Garufi:1937ja}
\bibfield{author}{\bibinfo{person}{Carlo~Alberto Garufi}.}
  \bibinfo{year}{1937}\natexlab{}.
\newblock \showarticletitle{{Ryccardi de Sancto Germano notarii Chronica}}.
\newblock \bibinfo{journal}{\emph{Rerum Italicarum Scriptores}}
  \bibinfo{volume}{7}, \bibinfo{number}{2} (\bibinfo{year}{1937}).
\newblock
\newblock
\shownote{{\url{https://bit.ly/2HHrE1W} (Archivio della Latinità Italiana del
  Medioevo), accessed 2018-05-28}.}


\bibitem[\protect\citeauthoryear{Gaudenzi}{Gaudenzi}{1971a}]%
        {Gaudenzi:1971hb}
\bibfield{author}{\bibinfo{person}{Augusto Gaudenzi}.}
  \bibinfo{year}{1971}\natexlab{a}.
\newblock \bibinfo{booktitle}{\emph{{Guido Faba - Dictamina Rhetorica
  Epistole}}}.
\newblock \bibinfo{publisher}{Forni Editore}, \bibinfo{address}{Bologna, IT},
  Chapter ``Guidonis Fabe Epistole’’.
\newblock
\newblock
\shownote{{\url{https://bit.ly/2WrxgWh} (Archivio della Latinità Italiana del
  Medioevo), accessed 2018-05-28}.}


\bibitem[\protect\citeauthoryear{Gaudenzi}{Gaudenzi}{1971b}]%
        {Gaudenzi:1971qu}
\bibfield{author}{\bibinfo{person}{Augusto Gaudenzi}.}
  \bibinfo{year}{1971}\natexlab{b}.
\newblock \bibinfo{booktitle}{\emph{{Guido Faba - Dictamina Rhetorica
  Epistole}}}.
\newblock \bibinfo{publisher}{Forni Editore}, \bibinfo{address}{Bologna, IT},
  Chapter ``Guidonis Fabe Dictamina Rhetorica’’.
\newblock
\newblock
\shownote{{\url{https://bit.ly/2I36vhI} (Archivio della Latinità Italiana del
  Medioevo), accessed 2018-05-28}.}


\bibitem[\protect\citeauthoryear{Griffin}{Griffin}{1936}]%
        {Griffin:1936ka}
\bibfield{author}{\bibinfo{person}{Nathaniel~E. Griffin}.}
  \bibinfo{year}{1936}\natexlab{}.
\newblock \bibinfo{booktitle}{\emph{{Guido De Columnis - Historia destructionis
  Troiae}}}.
\newblock \bibinfo{publisher}{The Mediaeval Academy of America},
  \bibinfo{address}{Cambridge, US}.
\newblock
\newblock
\shownote{{\url{https://bit.ly/2EF02IR} (Archivio della Latinità Italiana del
  Medioevo), accessed 2018-05-28}.}


\bibitem[\protect\citeauthoryear{Kristeller}{Kristeller}{1961}]%
        {Kristeller:1961la}
\bibfield{author}{\bibinfo{person}{Paul~O. Kristeller}.}
  \bibinfo{year}{1961}\natexlab{}.
\newblock \showarticletitle{{Un' `Ars dictaminis di Giovanni del Virgilio'}}.
\newblock \bibinfo{journal}{\emph{Italia Medioevale e Umanistica}}
  \bibinfo{number}{4} (\bibinfo{year}{1961}), \bibinfo{pages}{179--200}.
\newblock
\newblock
\shownote{{\url{https://bit.ly/2MeCD7k} (Archivio della Latinità Italiana del
  Medioevo), accessed 2018-05-28}.}


\bibitem[\protect\citeauthoryear{Mancuso}{Mancuso}{2015}]%
        {Mancuso:2015tz}
\bibfield{author}{\bibinfo{person}{Selene Mancuso}.}
  \bibinfo{year}{2015}\natexlab{}.
\newblock \emph{\bibinfo{title}{{Benvenuto da Imola, Glose Bucolicorum Virgilii
  (ad Buc. I, IV, VI, X). Studi critici ed edizione}}}.
\newblock \bibinfo{thesistype}{Master's\ thesis}. \bibinfo{school}{Department
  of Philology, Literature and Linguistics, University of Pisa},
  \bibinfo{address}{Pisa, IT}.
\newblock


\bibitem[\protect\citeauthoryear{Meloni}{Meloni}{1953}]%
        {Meloni:1953rb}
\bibfield{author}{\bibinfo{person}{Pietro Meloni}.}
  \bibinfo{year}{1953}\natexlab{}.
\newblock \bibinfo{booktitle}{\emph{{Nicolai Treveti Expositio L. Annaei
  Senecae Agamemnonis}}}.
\newblock \bibinfo{publisher}{Centro di Studi Filologici e Linguistici
  Siciliani}, \bibinfo{address}{Palermo, IT}.
\newblock
\newblock
\shownote{{\url{https://bit.ly/2KfRxYt} (Biblioteca Italiana), accessed
  2018-05-28}.}


\bibitem[\protect\citeauthoryear{Menestò and Brufani}{Menestò and
  Brufani}{1995}]%
        {Menesto:1995hz}
\bibfield{author}{\bibinfo{person}{Enrico Menestò} {and}
  \bibinfo{person}{Stefano Brufani}.} \bibinfo{year}{1995}\natexlab{}.
\newblock \bibinfo{booktitle}{\emph{{Fontes Franciscani}}}.
\newblock \bibinfo{publisher}{Edizioni Porziuncola}, \bibinfo{address}{Assisi,
  IT}.
\newblock


\bibitem[\protect\citeauthoryear{Monleone}{Monleone}{1941}]%
        {Monleone:1941ua}
\bibfield{author}{\bibinfo{person}{Giovanni Monleone}.}
  \bibinfo{year}{1941}\natexlab{}.
\newblock \bibinfo{booktitle}{\emph{{Jacopo da Varagine e la sua Cronaca di
  Genova dalle origini al MCCXCVII}}}.
\newblock \bibinfo{publisher}{FSI}, \bibinfo{address}{Roma, IT},
  \bibinfo{pages}{3--414}.
\newblock
\newblock
\shownote{{\url{https://bit.ly/2MhKlxs} (Archivio della Latinità Italiana del
  Medioevo), accessed 2018-05-28}.}


\bibitem[\protect\citeauthoryear{Nardi}{Nardi}{1996}]%
        {Nardi:1996fd}
\bibfield{author}{\bibinfo{person}{Bruno Nardi}.}
  \bibinfo{year}{1996}\natexlab{}.
\newblock \bibinfo{booktitle}{\emph{{Dante Alighieri - Opere minori}}}.
\newblock \bibinfo{publisher}{Riccardo Ricciardi Editore},
  \bibinfo{address}{Milano, IT}, Chapter ``Monarchia’’.
\newblock
\newblock
\shownote{{\url{https://bit.ly/2EEGQLg} (Biblioteca Italiana), accessed
  2018-05-28}.}


\bibitem[\protect\citeauthoryear{Romano}{Romano}{2004}]%
        {Romano:2004ap}
\bibfield{author}{\bibinfo{person}{Marta M.~M. Romano}.}
  \bibinfo{year}{2004}\natexlab{}.
\newblock \bibinfo{booktitle}{\emph{{Corpus Christianorum Continuatio
  Mediaevalis CLXXXIII}}}.
\newblock \bibinfo{publisher}{Brepols Publisher}, \bibinfo{address}{Turnhout,
  BE}, Chapter ``Raimundus Lullus -- Ars amativa boni’’,
  \bibinfo{pages}{120--432}.
\newblock
\newblock
\shownote{{\url{https://bit.ly/2wnBu2t} (Archivio della Latinità Italiana del
  Medioevo), accessed 2018-05-28}.}


\bibitem[\protect\citeauthoryear{Romano}{Romano}{1951}]%
        {Romano:1951uz}
\bibfield{author}{\bibinfo{person}{Vincenzo Romano}.}
  \bibinfo{year}{1951}\natexlab{}.
\newblock \bibinfo{booktitle}{\emph{{Giovanni Boccaccio - Genealogie deorum
  gentilium libri}}}.
\newblock \bibinfo{publisher}{Laterza}, \bibinfo{address}{Bari, IT}.
\newblock
\newblock
\shownote{{\url{https://bit.ly/2ZEvcZI} (Biblioteca Italiana), accessed
  2018-05-28}.}


\bibitem[\protect\citeauthoryear{Rossi}{Rossi}{1998}]%
        {Rossi:1999hq}
\bibfield{author}{\bibinfo{person}{Luca~C. Rossi}.}
  \bibinfo{year}{1998}\natexlab{}.
\newblock \bibinfo{booktitle}{\emph{Commento all' `Inferno' di Dante}}.
\newblock \bibinfo{publisher}{Scuola Normale Superiore},
  \bibinfo{address}{Pisa, IT}.
\newblock
\newblock
\shownote{Reprinted in Rossi, Luca C. (ed.), ``{I commenti danteschi dei secoli
  XIV, XV e XVI}'', LEXIS Progetti Editoriali, Roma, IT, 1999
  {\url{https://bit.ly/2EzazoX} (Biblioteca Italiana), accessed 2018-05-28}.}


\bibitem[\protect\citeauthoryear{Rossi}{Rossi}{2002}]%
        {Rossi:2002ip}
\bibfield{author}{\bibinfo{person}{Luca~Carlo Rossi}.}
  \bibinfo{year}{2002}\natexlab{}.
\newblock \showarticletitle{{`Benevenutus de Ymola super Valerio Maximo'.
  Ricerca sull'Expositio}}.
\newblock \bibinfo{journal}{\emph{Aevum - Rassegna di Scienze Storiche
  Linguistiche e Filologiche}} \bibinfo{number}{76} (\bibinfo{year}{2002}),
  \bibinfo{pages}{369--423}.
\newblock


\bibitem[\protect\citeauthoryear{Schneider}{Schneider}{1926}]%
        {Schneider:1926jw}
\bibfield{author}{\bibinfo{person}{Fedor Schneider}.}
  \bibinfo{year}{1926}\natexlab{}.
\newblock \showarticletitle{{Untersuchungen zur italienischen
  Verfassungsgeschichte: Staufisches aus der Formelsammlung des Petrus de
  Boateriis}}.
\newblock \bibinfo{journal}{\emph{Quellen und Forschungen aus italienischen
  Archiven und Bibliotheken}} \bibinfo{number}{18} (\bibinfo{year}{1926}),
  \bibinfo{pages}{191--273}.
\newblock


\bibitem[\protect\citeauthoryear{Stok}{Stok}{1991}]%
        {Stok:1991ft}
\bibfield{author}{\bibinfo{person}{Fabio Stok}.}
  \bibinfo{year}{1991}\natexlab{}.
\newblock \showarticletitle{{La `Vita di Virgilio' di Zono de' Magnalis}}.
\newblock \bibinfo{journal}{\emph{Rivista di Cultura Classica e Medioevale}}
  \bibinfo{volume}{33}, \bibinfo{number}{2} (\bibinfo{year}{1991}),
  \bibinfo{pages}{143--181}.
\newblock


\bibitem[\protect\citeauthoryear{Sutter}{Sutter}{1894}]%
        {Sutter:1894rv}
\bibfield{author}{\bibinfo{person}{Carl Sutter}.}
  \bibinfo{year}{1894}\natexlab{}.
\newblock \bibinfo{booktitle}{\emph{{Aus Leben und Schriften des Magisters
  Boncompagno}}}.
\newblock \bibinfo{publisher}{Akademische Verlagsbuchhandlung von J.C.B. Mohr},
  \bibinfo{address}{Freiburg im Breisgau, DE}.
\newblock
\newblock
\shownote{{\url{https://bit.ly/2ws07eg} (Archivio della Latinità Italiana del
  Medioevo), accessed 2018-05-28}.}


\bibitem[\protect\citeauthoryear{Tavoni}{Tavoni}{2011}]%
        {Tavoni:2011by}
\bibfield{author}{\bibinfo{person}{Mirko Tavoni}.}
  \bibinfo{year}{2011}\natexlab{}.
\newblock \bibinfo{booktitle}{\emph{{Dante Alighieri - Opere}}}.
\newblock \bibinfo{publisher}{Mondadori}, \bibinfo{address}{Milano, IT},
  Chapter ``De vulgari eloquentia’’.
\newblock
\newblock
\shownote{{\url{https://bit.ly/2HIfBRW} (DanteSearch), accessed 2018-05-28}.}


\bibitem[\protect\citeauthoryear{Ussani}{Ussani}{1959}]%
        {Ussani:1959tg}
\bibfield{author}{\bibinfo{person}{Vincenzo Ussani, Jr}.}
  \bibinfo{year}{1959}\natexlab{}.
\newblock \bibinfo{booktitle}{\emph{{L. Annaei Senecae Hercules furens et
  Nicolai Treveti expositio}}}.
\newblock \bibinfo{publisher}{Edizioni dell'Ateneo}, \bibinfo{address}{Roma,
  IT}.
\newblock
\newblock
\shownote{{\url{https://bit.ly/2KccRxN} (Biblioteca Italiana), accessed
  2018-05-28}.}


\bibitem[\protect\citeauthoryear{Zaccaria}{Zaccaria}{1967}]%
        {Zaccaria:1967oa}
\bibfield{author}{\bibinfo{person}{Vittorio Zaccaria}.}
  \bibinfo{year}{1967}\natexlab{}.
\newblock \bibinfo{booktitle}{\emph{Tutte le opere di {Giovanni Boccaccio}}}.
\newblock \bibinfo{publisher}{Mondadori}, \bibinfo{address}{Milano, IT},
  Chapter ``De mulieribus claris’’.
\newblock
\newblock
\shownote{{\url{https://bit.ly/2I1VWvl} (Biblioteca Italiana), accessed
  2018-05-28}.}


\end{thebibliography}

\begin{thebibliography}{30}


\ifx \showCODEN    \undefined \def \showCODEN     #1{\unskip}     \fi
\ifx \showDOI      \undefined \def \showDOI       #1{#1}\fi
\ifx \showISBNx    \undefined \def \showISBNx     #1{\unskip}     \fi
\ifx \showISBNxiii \undefined \def \showISBNxiii  #1{\unskip}     \fi
\ifx \showISSN     \undefined \def \showISSN      #1{\unskip}     \fi
\ifx \showLCCN     \undefined \def \showLCCN      #1{\unskip}     \fi
\ifx \shownote     \undefined \def \shownote      #1{#1}          \fi
\ifx \showarticletitle \undefined \def \showarticletitle #1{#1}   \fi
\ifx \showURL      \undefined \def \showURL       {\relax}        \fi
\providecommand\bibfield[2]{#2}
\providecommand\bibinfo[2]{#2}
\providecommand\natexlab[1]{#1}
\providecommand\showeprint[2][]{arXiv:#2}

\bibitem[\protect\citeauthoryear{Aggarwal and Zhai}{Aggarwal and Zhai}{2012}]%
        {Aggarwal:2012wl}
\bibfield{author}{\bibinfo{person}{Charu~C. Aggarwal} {and}
  \bibinfo{person}{ChengXiang Zhai}.} \bibinfo{year}{2012}\natexlab{}.
\newblock \showarticletitle{A survey of text classification algorithms}.
\newblock In \bibinfo{booktitle}{\emph{Mining Text Data}},
  \bibfield{editor}{\bibinfo{person}{Charu~C. Aggarwal} {and}
  \bibinfo{person}{ChengXiang Zhai}} (Eds.). \bibinfo{publisher}{Springer},
  \bibinfo{address}{Heidelberg, {DE}}, \bibinfo{pages}{163--222}.
\newblock


\bibitem[\protect\citeauthoryear{Casadei}{Casadei}{2020}]%
        {Casadei:2020tm}
\bibfield{editor}{\bibinfo{person}{Alberto Casadei}} (Ed.).
  \bibinfo{year}{2020}\natexlab{}.
\newblock \bibinfo{booktitle}{\emph{Atti del Seminario ``Nuove Inchieste
  sull'Epistola a {Cangrande}''}}.
\newblock \bibinfo{publisher}{Pisa University Press}, \bibinfo{address}{Pisa,
  {IT}}.
\newblock


\bibitem[\protect\citeauthoryear{Chaski}{Chaski}{2005}]%
        {Chaski:2005pd}
\bibfield{author}{\bibinfo{person}{Carole~E. Chaski}.}
  \bibinfo{year}{2005}\natexlab{}.
\newblock \showarticletitle{Who's at the keyboard? {A}uthorship attribution in
  digital evidence investigations}.
\newblock \bibinfo{journal}{\emph{International Journal of Digital Evidence}}
  \bibinfo{volume}{4}, \bibinfo{number}{1} (\bibinfo{year}{2005}).
\newblock


\bibitem[\protect\citeauthoryear{Corbara}{Corbara}{2019}]%
        {Corbara:2019rw}
\bibfield{author}{\bibinfo{person}{Silvia Corbara}.}
  \bibinfo{year}{2019}\natexlab{}.
\newblock \emph{\bibinfo{title}{The {Epistle to Cangrande} through the lens of
  computational authorship verification}}.
\newblock \bibinfo{thesistype}{Master's\ thesis}. \bibinfo{school}{Department
  of Philology, Literature, and Linguistics, University of Pisa},
  \bibinfo{address}{Pisa, {IT}}.
\newblock


\bibitem[\protect\citeauthoryear{Corbara, Moreo, Sebastiani, and
  Tavoni}{Corbara et~al\mbox{.}}{2019}]%
        {Corbara:2019cq}
\bibfield{author}{\bibinfo{person}{Silvia Corbara}, \bibinfo{person}{Alejandro
  Moreo}, \bibinfo{person}{Fabrizio Sebastiani}, {and} \bibinfo{person}{Mirko
  Tavoni}.} \bibinfo{year}{2019}\natexlab{}.
\newblock \showarticletitle{The {Epistle to Cangrande} through the lens of
  computational authorship verification}. In
  \bibinfo{booktitle}{\emph{Proceedings of the 1st International Workshop on
  Pattern Recognition for Cultural Heritage (PatReCH 2019)}}
  \emph{(\bibinfo{series}{Lecture Notes in Computer Science})}.
  \bibinfo{publisher}{Springer}, \bibinfo{address}{Trento, IT},
  \bibinfo{pages}{148--158}.
\newblock
\urldef\tempurl%
\url{https://doi.org/10.1007/978-3-030-30754-7_15}
\showDOI{\tempurl}


\bibitem[\protect\citeauthoryear{Forstall, Jacobson, and Scheirer}{Forstall
  et~al\mbox{.}}{2011}]%
        {Forstall:2011ev}
\bibfield{author}{\bibinfo{person}{Christopher~W. Forstall},
  \bibinfo{person}{Sarah~L. Jacobson}, {and} \bibinfo{person}{Walter~J.
  Scheirer}.} \bibinfo{year}{2011}\natexlab{}.
\newblock \showarticletitle{Evidence of intertextuality: {Investigating Paul
  the Deacon's Angustae Vitae}}.
\newblock \bibinfo{journal}{\emph{Literary and Linguistic Computing}}
  \bibinfo{volume}{26}, \bibinfo{number}{3} (\bibinfo{year}{2011}),
  \bibinfo{pages}{285--296}.
\newblock


\bibitem[\protect\citeauthoryear{Juola}{Juola}{2006}]%
        {Juola:2006fk}
\bibfield{author}{\bibinfo{person}{Patrick Juola}.}
  \bibinfo{year}{2006}\natexlab{}.
\newblock \showarticletitle{Authorship attribution}.
\newblock \bibinfo{journal}{\emph{Foundations and Trends in Information
  Retrieval}} \bibinfo{volume}{1}, \bibinfo{number}{3} (\bibinfo{year}{2006}),
  \bibinfo{pages}{233--334}.
\newblock
\urldef\tempurl%
\url{https://doi.org/10.1561/1500000005}
\showDOI{\tempurl}


\bibitem[\protect\citeauthoryear{Kabala}{Kabala}{2020}]%
        {Kabala:2020bu}
\bibfield{author}{\bibinfo{person}{Jakub Kabala}.}
  \bibinfo{year}{2020}\natexlab{}.
\newblock \showarticletitle{Computational authorship attribution in medieval
  {Latin corpora: The case of the Monk of Lido (ca. 1101--08) and Gallus
  Anonymous (ca. 1113--17)}}.
\newblock \bibinfo{journal}{\emph{Language Resources and Evaluation}}
  \bibinfo{volume}{54}, \bibinfo{number}{1} (\bibinfo{year}{2020}),
  \bibinfo{pages}{25–--56}.
\newblock
\urldef\tempurl%
\url{https://doi.org/10.1007/s10579-018-9424-0}
\showDOI{\tempurl}


\bibitem[\protect\citeauthoryear{Kestemont, Moens, and Deploige}{Kestemont
  et~al\mbox{.}}{2015}]%
        {Kestemont:2015lp}
\bibfield{author}{\bibinfo{person}{Mike Kestemont}, \bibinfo{person}{Sara
  Moens}, {and} \bibinfo{person}{Jeroen Deploige}.}
  \bibinfo{year}{2015}\natexlab{}.
\newblock \showarticletitle{Collaborative authorship in the twelfth century:
  {A} stylometric study of {Hildegard of Bingen and Guibert of Gembloux}}.
\newblock \bibinfo{journal}{\emph{Digital Scholarship in the Humanities}}
  \bibinfo{volume}{30}, \bibinfo{number}{2} (\bibinfo{year}{2015}),
  \bibinfo{pages}{199--224}.
\newblock
\urldef\tempurl%
\url{https://doi.org/10.1093/llc/fqt063}
\showDOI{\tempurl}


\bibitem[\protect\citeauthoryear{Kestemont, Stover, Koppel, Karsdorp, and
  Daelemans}{Kestemont et~al\mbox{.}}{2016}]%
        {Kestemont:2016fh}
\bibfield{author}{\bibinfo{person}{Mike Kestemont}, \bibinfo{person}{Justin~A.
  Stover}, \bibinfo{person}{Moshe Koppel}, \bibinfo{person}{Folgert Karsdorp},
  {and} \bibinfo{person}{Walter Daelemans}.} \bibinfo{year}{2016}\natexlab{}.
\newblock \showarticletitle{Authenticating the writings of {Julius Caesar}}.
\newblock \bibinfo{journal}{\emph{Expert Systems with Applications}}
  \bibinfo{volume}{63} (\bibinfo{year}{2016}), \bibinfo{pages}{86--96}.
\newblock
\urldef\tempurl%
\url{https://doi.org/10.1016/j.eswa.2016.06.029}
\showDOI{\tempurl}


\bibitem[\protect\citeauthoryear{Koppel, Argamon, and Shimoni}{Koppel
  et~al\mbox{.}}{2002}]%
        {Koppel02}
\bibfield{author}{\bibinfo{person}{Moshe Koppel}, \bibinfo{person}{Shlomo
  Argamon}, {and} \bibinfo{person}{Anat~R. Shimoni}.}
  \bibinfo{year}{2002}\natexlab{}.
\newblock \showarticletitle{Automatically categorizing written texts by author
  gender}.
\newblock \bibinfo{journal}{\emph{Literary and Linguistic Computing}}
  \bibinfo{volume}{17}, \bibinfo{number}{4} (\bibinfo{year}{2002}),
  \bibinfo{pages}{401--412}.
\newblock
\urldef\tempurl%
\url{https://doi.org/10.1093/llc/17.4.401}
\showDOI{\tempurl}


\bibitem[\protect\citeauthoryear{Koppel and Schler}{Koppel and Schler}{2004}]%
        {Koppel:2004ja}
\bibfield{author}{\bibinfo{person}{Moshe Koppel} {and}
  \bibinfo{person}{Jonathan Schler}.} \bibinfo{year}{2004}\natexlab{}.
\newblock \showarticletitle{Authorship verification as a one-class
  classification problem}. In \bibinfo{booktitle}{\emph{Proceedings of the 21st
  International Conference on Machine Learning (ICML 2004)}}.
  \bibinfo{address}{Banff, CA}.
\newblock
\urldef\tempurl%
\url{https://doi.org/10.1145/1015330.1015448}
\showDOI{\tempurl}


\bibitem[\protect\citeauthoryear{Koppel, Schler, and Argamon}{Koppel
  et~al\mbox{.}}{2009}]%
        {Koppel:2009ix}
\bibfield{author}{\bibinfo{person}{Moshe Koppel}, \bibinfo{person}{Jonathan
  Schler}, {and} \bibinfo{person}{Shlomo Argamon}.}
  \bibinfo{year}{2009}\natexlab{}.
\newblock \showarticletitle{Computational methods in authorship attribution}.
\newblock \bibinfo{journal}{\emph{Journal of the American Society for
  Information Science and Technology}} \bibinfo{volume}{60},
  \bibinfo{number}{1} (\bibinfo{year}{2009}), \bibinfo{pages}{9--26}.
\newblock
\urldef\tempurl%
\url{https://doi.org/10.1002/asi.20961}
\showDOI{\tempurl}


\bibitem[\protect\citeauthoryear{Koppel and Winter}{Koppel and Winter}{2014}]%
        {Koppel:2014bq}
\bibfield{author}{\bibinfo{person}{Moshe Koppel} {and} \bibinfo{person}{Yaron
  Winter}.} \bibinfo{year}{2014}\natexlab{}.
\newblock \showarticletitle{Determining if two documents are written by the
  same author}.
\newblock \bibinfo{journal}{\emph{Journal of the Association for Information
  Science and Technology}} \bibinfo{volume}{65}, \bibinfo{number}{1}
  (\bibinfo{year}{2014}), \bibinfo{pages}{178--187}.
\newblock
\urldef\tempurl%
\url{https://doi.org/10.1002/asi.22954}
\showDOI{\tempurl}


\bibitem[\protect\citeauthoryear{Larner}{Larner}{2014}]%
        {Larner:2014kl}
\bibfield{author}{\bibinfo{person}{Samuel Larner}.}
  \bibinfo{year}{2014}\natexlab{}.
\newblock \bibinfo{booktitle}{\emph{Forensic Authorship Analysis and the {World
  Wide Web}}}.
\newblock \bibinfo{publisher}{Springer}, \bibinfo{address}{Heidelberg, {DE}}.
\newblock


\bibitem[\protect\citeauthoryear{Niculescu-Mizil and Caruana}{Niculescu-Mizil
  and Caruana}{2005}]%
        {Niculescu-Mizil:2005gh}
\bibfield{author}{\bibinfo{person}{Alexandru Niculescu-Mizil} {and}
  \bibinfo{person}{Rich Caruana}.} \bibinfo{year}{2005}\natexlab{}.
\newblock \showarticletitle{Predicting good probabilities with supervised
  learning}. In \bibinfo{booktitle}{\emph{Proceedings of the 22nd International
  Conference on Machine Learning (ICML 2005)}}. \bibinfo{address}{Bonn, {DE}},
  \bibinfo{pages}{625--632}.
\newblock
\urldef\tempurl%
\url{https://doi.org/10.1145/1102351.1102430}
\showDOI{\tempurl}


\bibitem[\protect\citeauthoryear{Pellegrini}{Pellegrini}{2018}]%
        {Pellegrini:2018yb}
\bibfield{author}{\bibinfo{person}{Paolo Pellegrini}.}
  \bibinfo{year}{2018}\natexlab{}.
\newblock \showarticletitle{La quattordicesima epistola di Dante Alighieri:
  {P}rimi appunti per una attribuzione}.
\newblock \bibinfo{journal}{\emph{Studi di Erudizione e di Filologia Italiana}}
   \bibinfo{volume}{7} (\bibinfo{year}{2018}), \bibinfo{pages}{5--20}.
\newblock


\bibitem[\protect\citeauthoryear{Perkins}{Perkins}{2015}]%
        {Perkins:2015rp}
\bibfield{author}{\bibinfo{person}{Ria Perkins}.}
  \bibinfo{year}{2015}\natexlab{}.
\newblock \showarticletitle{Native language identification {(NLID)} for
  forensic authorship analysis of weblogs}.
\newblock In \bibinfo{booktitle}{\emph{New Threats and Countermeasures in
  Digital Crime and Cyber Terrorism}},
  \bibfield{editor}{\bibinfo{person}{Maurice Dawson} {and}
  \bibinfo{person}{Marwan Omar}} (Eds.). \bibinfo{publisher}{IGI Global},
  \bibinfo{address}{Hershey, {US}}, \bibinfo{pages}{213--234}.
\newblock
\urldef\tempurl%
\url{https://doi.org/0.4018/978-1-4666-8345-7.ch012}
\showDOI{\tempurl}


\bibitem[\protect\citeauthoryear{Rocha, Scheirer, Forstall, Cavalcante,
  Theophilo, Shen, Carvalho, and Stamatatos}{Rocha et~al\mbox{.}}{2017}]%
        {Rocha:2017yy}
\bibfield{author}{\bibinfo{person}{Anderson Rocha}, \bibinfo{person}{Walter~J.
  Scheirer}, \bibinfo{person}{Christopher~W. Forstall}, \bibinfo{person}{Thiago
  Cavalcante}, \bibinfo{person}{Antonio Theophilo}, \bibinfo{person}{Bingyu
  Shen}, \bibinfo{person}{Ariadne Carvalho}, {and} \bibinfo{person}{Efstathios
  Stamatatos}.} \bibinfo{year}{2017}\natexlab{}.
\newblock \showarticletitle{Authorship attribution for social media forensics}.
\newblock \bibinfo{journal}{\emph{{IEEE} Transactions on Information Forensics
  and Security}} \bibinfo{volume}{12}, \bibinfo{number}{1}
  (\bibinfo{year}{2017}), \bibinfo{pages}{5--33}.
\newblock
\urldef\tempurl%
\url{https://doi.org/10.1109/TIFS.2016.2603960}
\showDOI{\tempurl}


\bibitem[\protect\citeauthoryear{Salton and Buckley}{Salton and
  Buckley}{1988}]%
        {Salton88}
\bibfield{author}{\bibinfo{person}{Gerard Salton} {and}
  \bibinfo{person}{Christopher Buckley}.} \bibinfo{year}{1988}\natexlab{}.
\newblock \showarticletitle{Term-weighting approaches in automatic text
  retrieval}.
\newblock \bibinfo{journal}{\emph{Information Processing and Management}}
  \bibinfo{volume}{24}, \bibinfo{number}{5} (\bibinfo{year}{1988}),
  \bibinfo{pages}{513--523}.
\newblock


\bibitem[\protect\citeauthoryear{Savoy}{Savoy}{2019}]%
        {Savoy:2019qr}
\bibfield{author}{\bibinfo{person}{Jacques Savoy}.}
  \bibinfo{year}{2019}\natexlab{}.
\newblock \showarticletitle{Authorship of Pauline epistles revisited}.
\newblock \bibinfo{journal}{\emph{Journal of the Association for Information
  Science and Technology}} \bibinfo{volume}{70}, \bibinfo{number}{10}
  (\bibinfo{year}{2019}), \bibinfo{pages}{1089--1097}.
\newblock
\urldef\tempurl%
\url{https://doi.org/10.1002/asi.24176}
\showDOI{\tempurl}


\bibitem[\protect\citeauthoryear{Schmid, Iqbal, and Fung}{Schmid
  et~al\mbox{.}}{2015}]%
        {Schmid:2015qq}
\bibfield{author}{\bibinfo{person}{Michael~R. Schmid},
  \bibinfo{person}{Farkhund Iqbal}, {and} \bibinfo{person}{Benjamin C.~M.
  Fung}.} \bibinfo{year}{2015}\natexlab{}.
\newblock \showarticletitle{E-mail authorship attribution using customized
  associative classification}.
\newblock \bibinfo{journal}{\emph{Digital Investigation}} \bibinfo{volume}{14},
  \bibinfo{number}{1} (\bibinfo{year}{2015}), \bibinfo{pages}{S116--S126}.
\newblock
\urldef\tempurl%
\url{https://doi.org/10.1016/j.diin.2015.05.012}
\showDOI{\tempurl}


\bibitem[\protect\citeauthoryear{Sebastiani}{Sebastiani}{2002}]%
        {ACMCS02}
\bibfield{author}{\bibinfo{person}{Fabrizio Sebastiani}.}
  \bibinfo{year}{2002}\natexlab{}.
\newblock \showarticletitle{Machine learning in automated text categorization}.
\newblock \bibinfo{journal}{\emph{Comput. Surveys}} \bibinfo{volume}{34},
  \bibinfo{number}{1} (\bibinfo{year}{2002}), \bibinfo{pages}{1--47}.
\newblock
\urldef\tempurl%
\url{https://doi.org/10.1145/505282.505283}
\showDOI{\tempurl}


\bibitem[\protect\citeauthoryear{Stamatatos}{Stamatatos}{2009}]%
        {Stamatatos2009:yq}
\bibfield{author}{\bibinfo{person}{Efstathios Stamatatos}.}
  \bibinfo{year}{2009}\natexlab{}.
\newblock \showarticletitle{A survey of modern authorship attribution methods}.
\newblock \bibinfo{journal}{\emph{Journal of the American Society for
  Information Science and Technology}} \bibinfo{volume}{60},
  \bibinfo{number}{3} (\bibinfo{year}{2009}), \bibinfo{pages}{538--556}.
\newblock
\urldef\tempurl%
\url{https://doi.org/10.1002/asi.21001}
\showDOI{\tempurl}


\bibitem[\protect\citeauthoryear{Stamatatos}{Stamatatos}{2016}]%
        {Stamatatos:2016ij}
\bibfield{author}{\bibinfo{person}{Efstathios Stamatatos}.}
  \bibinfo{year}{2016}\natexlab{}.
\newblock \showarticletitle{Authorship verification: {A} review of recent
  advances}.
\newblock \bibinfo{journal}{\emph{Research in Computing Science}}
  \bibinfo{volume}{123} (\bibinfo{year}{2016}), \bibinfo{pages}{9--25}.
\newblock


\bibitem[\protect\citeauthoryear{Stover, Winter, Koppel, and Kestemont}{Stover
  et~al\mbox{.}}{2016}]%
        {Stover:2016zl}
\bibfield{author}{\bibinfo{person}{Justin~A. Stover}, \bibinfo{person}{Yaron
  Winter}, \bibinfo{person}{Moshe Koppel}, {and} \bibinfo{person}{Mike
  Kestemont}.} \bibinfo{year}{2016}\natexlab{}.
\newblock \showarticletitle{Computational authorship verification method
  attributes a new work to a major 2nd century {African} author}.
\newblock \bibinfo{journal}{\emph{Journal of the American Society for
  Information Science and Technology}} \bibinfo{volume}{67},
  \bibinfo{number}{1} (\bibinfo{year}{2016}), \bibinfo{pages}{239--242}.
\newblock
\urldef\tempurl%
\url{https://doi.org/10.1002/asi.23460}
\showDOI{\tempurl}


\bibitem[\protect\citeauthoryear{Toynbee}{Toynbee}{1918}]%
        {Toynbee:1918mj}
\bibfield{author}{\bibinfo{person}{Paget Toynbee}.}
  \bibinfo{year}{1918}\natexlab{}.
\newblock \showarticletitle{Dante and the ``cursus'': {A} new argument in
  favour of the authenticity of the {Quaestio de Aqua et Terra}}.
\newblock \bibinfo{journal}{\emph{The Modern Language Review}}
  \bibinfo{volume}{13}, \bibinfo{number}{4} (\bibinfo{year}{1918}),
  \bibinfo{pages}{420--430}.
\newblock


\bibitem[\protect\citeauthoryear{Tuccinardi}{Tuccinardi}{2017}]%
        {Tuccinardi:2017yg}
\bibfield{author}{\bibinfo{person}{Enrico Tuccinardi}.}
  \bibinfo{year}{2017}\natexlab{}.
\newblock \showarticletitle{An application of a profile-based method for
  authorship verification: {I}nvestigating the authenticity of {Pliny the
  Younger's letter to Trajan concerning the Christians}}.
\newblock \bibinfo{journal}{\emph{Digital Scholarship in the Humanities}}
  \bibinfo{volume}{32}, \bibinfo{number}{2} (\bibinfo{year}{2017}),
  \bibinfo{pages}{435--447}.
\newblock
\urldef\tempurl%
\url{https://doi.org/10.1093/llc/fqw001}
\showDOI{\tempurl}


\bibitem[\protect\citeauthoryear{Vainio, Välimäki, Hella, Kaartinen, Immonen,
  Vesanto, and Ginter}{Vainio et~al\mbox{.}}{2019}]%
        {Vainio:2019cg}
\bibfield{author}{\bibinfo{person}{Raija Vainio}, \bibinfo{person}{Reima
  Välimäki}, \bibinfo{person}{Anni Hella}, \bibinfo{person}{Marjo Kaartinen},
  \bibinfo{person}{Teemu Immonen}, \bibinfo{person}{Aleksi Vesanto}, {and}
  \bibinfo{person}{Filip Ginter}.} \bibinfo{year}{2019}\natexlab{}.
\newblock \showarticletitle{Reconsidering authorship in the {Ciceronian} corpus
  through computational authorship attribution}.
\newblock \bibinfo{journal}{\emph{Ciceroniana On Line}} \bibinfo{volume}{3},
  \bibinfo{number}{1} (\bibinfo{year}{2019}).
\newblock


\bibitem[\protect\citeauthoryear{Yang and Pedersen}{Yang and Pedersen}{1997}]%
        {Yang97}
\bibfield{author}{\bibinfo{person}{Yiming Yang} {and} \bibinfo{person}{Jan~O.
  Pedersen}.} \bibinfo{year}{1997}\natexlab{}.
\newblock \showarticletitle{A comparative study on feature selection in text
  categorization}. In \bibinfo{booktitle}{\emph{Proceedings of the 14th
  International Conference on Machine Learning (ICML 1997)}}.
  \bibinfo{address}{Nashville, {US}}, \bibinfo{pages}{412--420}.
\newblock


\end{thebibliography}
\end{document}